\documentclass{article}

\usepackage{iclr2027_conference,times}
\usepackage{memory_arxiv}
\usepackage{graphicx}
\usepackage{booktabs}
\usepackage{amsmath,amssymb}
\usepackage{microtype}
\usepackage{url}
\usepackage{enumitem}
\usepackage{placeins}
\usepackage{float}
\usepackage{xcolor}
\usepackage{amsthm}
\usepackage{hyperref}
\hypersetup{hidelinks,pdftitle={When Does a Memory Descendant Need Repair? Structural Lineage vs. Stale State in Persistent Agent Memory},pdfauthor={Joongwon Chae, Lihui Luo, Wenjie Li, Yang Liu, Zichen Liang, Peter E. Lobie, Peiwu Qin}}
\newtheorem{proposition}{Proposition}

\title{When Does a Memory Descendant Need Repair?\\Structural Lineage vs. Stale State\\in Persistent Agent Memory}

\author{Joongwon Chae\textsuperscript{1,\#}\quad Lihui Luo\textsuperscript{1,\#}\quad Wenjie Li\textsuperscript{2}\quad Yang Liu\textsuperscript{1}\\Zichen Liang\textsuperscript{3}\quad Peter E. Lobie\textsuperscript{1}\quad Peiwu Qin\textsuperscript{4,*}}
\affiliations{\textsuperscript{1}Tsinghua University\quad \textsuperscript{2}Shanghai Innovation Institute\\\textsuperscript{3}ByteDance Inc.\quad \textsuperscript{4}Chinese Medicine Guangdong Laboratory\\[3pt]\textsuperscript{\#}Equal contribution.\quad \textsuperscript{*}Corresponding author.}

\newcommand{\Gtrue}{G^{\star}}
\newcommand{\Gobs}{\widehat{G}}
\newcommand{\Astruct}{\mathcal{A}_{\mathrm{struct}}}
\newcommand{\Asens}{\mathcal{A}_{\mathrm{sens}}}
\newcommand{\Astale}{\mathcal{A}_{\mathrm{stale}}}

\begin{document}
\maketitle

\begin{abstract}
Persistent LLM agents reuse intermediate conclusions stored across interactions. When a source fact is corrected, these conclusions may remain active and mislead later answers. Dependency-guided repair traces the source's structural descendants, but does reliable repair require recording the full lineage? We distinguish historical dependency, update sensitivity, and stored-state error, then compare provenance graphs under a fixed repair operator in controlled memory lifecycles and public multi-hop tasks. In the controlled testbed, LLM material-support provenance reaches $95$--$100\%$ of stale state despite covering only about half of structural descendants at greater depth. Matched-random graphs with identical edge budgets, precision, and recall reach only about two-thirds. Value-agreement support reproduces this selectivity, and closed-loop runs preserve the result when repairs regenerate answers and dependencies. A \emph{certificate}---a subset of parent values sufficient to force the stored conclusion---explains when edges can be omitted. We prove consistency across sequential and simultaneous corrections on fixed dependency graphs under local correctness, certificate containment, and full-parent re-derivation. On public lookup chains, LLM attribution achieves post-repair accuracy within $0.02$ of oracle path repair, while structural and stale-state coverage nearly coincide. Structural fidelity and repair adequacy therefore differ, with the available repair savings depending on the support needed to preserve each conclusion.
\end{abstract}

\section{Introduction}
\label{sec:intro}

Persistent LLM agents increasingly reuse conclusions stored in memory instead of recomputing every decision from raw history \citep{park2023generative,shinn2023reflexion,packer2023memgpt,xu2025amem}. Storing conclusions, however, creates a maintenance problem when upstream facts change: correcting an erroneous source fact (\emph{root}) does not automatically remove or update a downstream conclusion (\emph{derived memory}) written earlier. In our lifecycle experiments, correcting a source fact while leaving its outdated derived conclusion in memory drops post-correction answer accuracy from $0.996$ to $0.907$, because the outdated conclusion continues to be retrieved and misleads subsequent reasoning.

\begin{figure}[t]
  \centering
  \includegraphics[width=\linewidth]{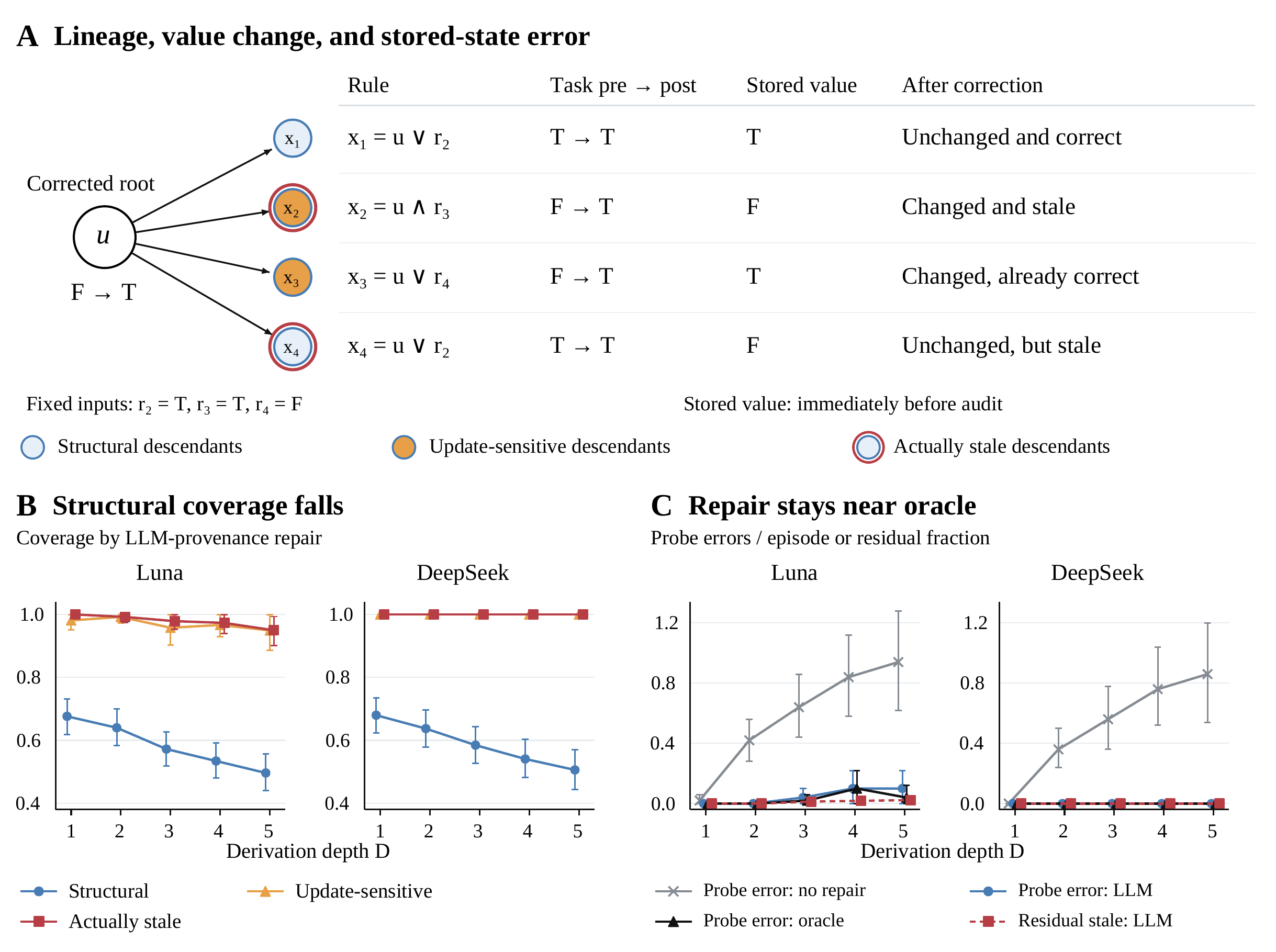}
  \caption{\textbf{Structural lineage, update sensitivity, and stored-state staleness are distinct repair targets.} \textbf{A:} When a root fact $u$ is corrected, its structural descendants include nodes whose task value does not change ($x_1$), nodes whose task value flips and become stale ($x_2$), and reader-error cases that are sensitive-but-fresh ($x_3$) or insensitive-but-stale ($x_4$). \textbf{B:} Under LLM material-support provenance, structural coverage falls with derivation depth $D$, while update-sensitive and stale-state coverage remain near $100\%$. \textbf{C:} End-of-lifecycle outcomes track stale-state coverage: residual stale memories stay near zero and post-repair answer error remains close to oracle repair. Error bars show episode-cluster paired bootstrap 95\% CIs across 50 episodes.}
  \label{fig:three_objects}
  \vspace{-4pt}
\end{figure}

To prevent outdated conclusions from corrupting future answers, a memory system must trace a \emph{provenance graph} from the corrected source fact to invalidate and re-derive affected descendants \citep{zhao2026memorepair,yu2026rollback}. Recording complete \emph{structural lineage}---every parent memory required by a task rule at formation---is sufficient, but changing a consulted parent does not always change a downstream conclusion: if a disjunctive rule $A\lor B$ held because both $A$ and $B$ were true, correcting $A$ to false while $B$ stays true leaves YES valid. Conversely, if the agent erred at formation, its stored answer may need repair even when the ground-truth answer is unchanged. Repair need thus depends on \emph{whether the correction changes the ground-truth answer} and \emph{what answer is currently stored}. For each corrected root $u$ at audit time $t$, we therefore separate three downstream sets (Fig.~\ref{fig:three_objects}A): (1) \emph{structural descendants} ($\Astruct$), which depended on $u$ when formed; (2) \emph{update-sensitive descendants} ($\Asens$), whose ground-truth task answer flips under the current correction; and (3) \emph{stale stored descendants} ($\Astale$), whose currently stored value disagrees with the post-correction truth and actually needs repair (only about one-quarter of $\Astruct$ in our testbed).

This distinction changes how we judge provenance graphs. As reasoning depth ($D$) grows from $1$ to $5$, an LLM prompted to record only candidate memories that materially support each conclusion captures only half ($50$--$51\%$) of structural descendants, yet reaches $95$--$100\%$ of stale stored state and achieves near-oracle post-repair accuracy (Fig.~\ref{fig:three_objects}B--C). A random graph matched to the exact same edge count, precision, and recall repairs only $66$--$68\%$ of stale state, and a single-parent counterfactual (\emph{but-for}) graph with perfect precision misses stale memories under overdetermined support.

Why does half the lineage suffice? Because the LLM nearly always retains a \emph{certificate}---a parent subset sufficient on its own to force the stored conclusion ($99.8$--$100\%$ of locally correct formations). A symbolic \emph{value-agreement} rule, keeping only parents whose truth value aligns with the stored conclusion, closely reproduces the LLM graph's repair profile. Formally, we prove that on unate rules value agreement reaches all update-sensitive descendants under single-root updates (Proposition~\ref{prop:vc}), and that for general deterministic rules over a fixed dependency DAG, local calculation correctness, certificate containment, and full-parent topological re-derivation keep all active memories consistent with current root facts across sequential and simultaneous corrections (Proposition~\ref{prop:certificate}). When a certificate is a proper subset of the structural parents, provenance can drop the remaining edges without losing repair adequacy. On public multi-hop lookups (MemoryAgentBench \citealt{hu2026memoryagentbench}), where updating an intermediate entity dynamically rewires the downstream lookup path and every active hop is essential, structural and stale coverage nearly coincide and LLM provenance matches oracle path repair.

Our main contributions are:
\begin{itemize}[leftmargin=1.5em,itemsep=1pt,topsep=2pt]
    \item \textbf{Separating structural lineage from repair need.} We formalize structural, update-sensitive, and stale stored descendants (Sec.~\ref{sec:problem}) and prove that certificate-containing provenance---including value agreement on unate rules---maintains consistency with current root facts across sequential and simultaneous corrections (Propositions~\ref{prop:vc}--\ref{prop:certificate}).
    \item \textbf{Controlled replay and closed-loop evaluation.} Depth sweeps and fixed-history controls (Secs.~\ref{sec:res_ablation}--\ref{sec:res_controls}) separate parent identity from edge count and structural fidelity. Closed-loop runs retain near-oracle accuracy with $44$--$57\%$ fewer re-derivations under LLM and certificate provenance (Sec.~\ref{sec:closed_loop}).
    \item \textbf{Public multi-hop validation.} On MemoryAgentBench FactConsolidation (Secs.~\ref{sec:public}, \ref{sec:public_llm}), we show that persistent intermediates harm accuracy specifically when invalidated by updates, confirm that structural and stale coverage nearly coincide on non-redundant lookup chains, and show that LLM provenance matches oracle repair.
\end{itemize}

\section{Related Work}
\label{sec:related}

\paragraph{Persistent memory and cascade repair in LLM agents.}
Agents retain observations, reflections, and intermediate conclusions across turns \citep{park2023generative,shinn2023reflexion,packer2023memgpt,xu2025amem}; when earlier facts are corrected, derived conclusions remain retrievable and bias answers unless invalidated \citep{shen2026revoked}. Recent systems address this via dependency-guided repair: \textsc{MemoRepair} withdraws reachable descendants and republishes a selected subset \citep{zhao2026memorepair}, \textsc{Revise} follows data and control dependencies during online revisions \citep{qi2026revise}, \textsc{Grade} logs execution dependencies \citep{zhao2026grade}, and \citet{yu2026rollback} roll back faulty steps while preserving nodes that retain an unaffected supporting parent. Existing repair policies and evaluation metrics, however, generally treat the full structural closure as the target: for instance, \textsc{MemoRepair} evaluates $\text{Leak\%}$ ($1 - C_{\mathrm{struct}}$ on active memories) and with default $\lambda=0.3$ keeps all $2{,}659/2{,}659$ reachable candidates in our testbed, matching oracle closure (Appendix~\ref{app:prior}). Judged by $\text{Leak\%}$, our LLM graph would be penalized for ``leaking'' $44$--$45\%$ of the structural cascade despite repairing $97$--$100\%$ of wrong memories. Conversely, the lenient support check of \citet{yu2026rollback} skips any node with one untouched parent, leaving conjunctive and threshold rules unrepaired (Table~\ref{tab:closed_loop}). Rather than changing the repair operator, we study which provenance edges allow reachability repair to fix all stale memories with minimal work.

\paragraph{Provenance, witnesses, and incremental maintenance.}
Database provenance characterizes which inputs witness a query result \citep{buneman2001why,green2007provenance,cheney2007dependency}, and sufficient reasons identify minimal feature subsets that guarantee a model's output \citep{ignatiev2019abduction,darwiche2020reasons}. For a correct output of a unate rule, value-agreement parents contain all minimal sufficient reasons (Appendix~\ref{app:robustness}); we extend this static concept to dynamic maintenance across sequential updates. Counterfactual causality introduces contingencies because single-factor (\emph{but-for}) tests fail under overdetermination \citep{halpern2005causes,meliou2010causality}, a failure mode our but-for baseline exposes when non-decisive parents become decisive after later updates. Truth-maintenance \citep{doyle1979truth,dekleer1986atms}, incremental view maintenance \citep{gupta1993views}, and self-adjusting computation \citep{acar2002adaptive,hammer2014adapton} propagate updates along recorded dependencies; we show how certificate selectivity cuts re-derivations without losing accuracy.

\section{Three Repair-Relevant Objects}
\label{sec:problem}

Consider a persistent memory store containing raw source facts (\emph{roots}) and LLM-generated intermediate conclusions (\emph{derived memories}). Each derived memory $v$ has a deterministic local rule $f_v$ over its direct structural parents $\mathrm{Pa}(v)$, inducing a ground-truth task function $F_v(X)$ over root assignment $X$, while $y_v$ denotes the answer stored by the reader LLM (notation in Appendix Table~\ref{tab:notation}). When a root $u$ is audited and corrected at time $t$, we distinguish three downstream sets.

\subsection{Structural descendants}
Let $\Gtrue_t$ be the oracle structural graph over memories active just before audit time $t$, where $a\rightarrow v$ means $a$ was a rule-required parent presented at formation. The \emph{structural descendants} of $u$ at time $t$ are all active memories connected to $u$ by a directed path in $\Gtrue_t$:
\begin{equation}
\Astruct(u,t)=\{v: u\leadsto v \text{ in } \Gtrue_t\}.
\end{equation}

\subsection{Update-sensitive descendants}
Let $X_{t^-}$ and $X_{t^+}$ be the root assignments just before and after auditing $u$ at time $t$ (including all earlier corrections). The \emph{update-sensitive descendants} are those whose ground-truth task answer flips under this update (evaluated at audit time $t$ rather than formation time because earlier corrections can change whether $u$ is decisive for $v$; Appendix~\ref{app:robustness}):
\begin{equation}
\Asens(u,t)=\{v\in\Astruct(u,t):F_v(X_{t^-})\neq F_v(X_{t^+})\}.
\end{equation}

\subsection{Stale stored descendants}
Because an LLM reader can err at formation, the active stored answer $y_v(t^-)$ just before audit $t$ may differ from $F_v(X_{t^-})$. Whether $v$ requires repair depends on whether its stored answer disagrees with the post-audit truth $F_v(X_{t^+})$. Among active structural descendants (which exclude memories already replaced by earlier repairs), the \emph{stale stored descendants} are:
\begin{equation}
\Astale(u,t)=\{v\in\Astruct(u,t):y_v(t^-)\neq F_v(X_{t^+})\}.
\end{equation}
As shown in Fig.~\ref{fig:three_objects}A, $\Asens$ and $\Astale$ need not be nested: a formation error can leave a stored answer wrong when $F_v$ is unchanged (\emph{insensitive-but-stale}, $x_4$), or matching the new truth after $F_v$ flips (\emph{sensitive-but-fresh}, $x_3$). Reader errors outside $u$'s lineage are tracked separately.

\paragraph{Coverage and lifecycle metrics.}
Given an observed provenance graph $\Gobs$, our repair operator invalidates the set $\widehat{A}(u,t)$ of active nodes reachable from $u$ in $\Gobs$ and re-derives them topologically. For $k\in\{\mathrm{struct},\mathrm{sens},\mathrm{stale}\}$, coverage is the pooled ratio of sums across all $(\text{correction},\text{descendant})$ events in a reader $\times$ depth cell:
\begin{equation}
C_k=\frac{\sum_{(u,t)}|\widehat{A}(u,t)\cap \mathcal{A}_k(u,t)|}{\sum_{(u,t)}|\mathcal{A}_k(u,t)|}.
\end{equation}
After all audits finish, we report the \emph{residual-stale fraction} (share of derived memories formed under an initially wrong ancestor whose stored value still disagrees with the final truth) and \emph{post-repair probe error} (wrong answers per episode on final write-back-free test queries).

\subsection{Why half the lineage can suffice: Value agreement and certificates}
\label{sec:adequacy}
Because repair follows paths in $\Gobs$, two graphs with identical edge counts, precision, and recall can differ sharply in repair adequacy depending on \emph{which} parents they retain. A rule $f_v$ is \emph{unate} if each parent enters with a fixed polarity, making $f_v$ monotone in each parent's signed truth value. Given state $X$, a parent of $v$ is in \emph{value agreement} with $v$ if its signed truth value matches $F_v(X)$ (e.g., true inputs of a true OR). Let $G^{\mathrm{va}}_X$ keep only value-agreement edges.

\begin{proposition}\label{prop:vc}
Let all local rules $f_v$ be unate, let $u$ be the only root whose value changes at audit $t$, and suppose every active stored value equals its task value under $X_{t^-}$. Then every $v\in\Asens(u,t)$ is reachable from $u$ in $G^{\mathrm{va}}_{X_{t^-}}$.
\end{proposition}

Intuitively, for a monotone rule's output to flip from YES to NO, at least one parent pushing the output toward YES (in value agreement with YES) must flip to NO; parents that were already NO only reinforce YES when flipped, so omitting them never misses a flip.

\paragraph{Certificates across sequential updates.}
Proposition~\ref{prop:vc} uses parent values just before an audit, whereas provenance edges are recorded earlier, at formation, before sequential corrections occur. Given parent values $x$ at formation, a parent subset $S\subseteq\mathrm{Pa}(v)$ is a \emph{certificate} of $v$ if fixing the parents in $S$ to $x_S$ forces the output $f_v(x)$ regardless of the remaining parents (minimal certificates are sufficient reasons \citep{darwiche2020reasons,ignatiev2019abduction}).

\begin{proposition}\label{prop:certificate}
Over a fixed structural DAG and fixed deterministic local rules $f_v$ over finite parent domains during the correction lifecycle, suppose that (i) each stored value is computed correctly from its currently stored structural parents when written, (ii) each formation records incoming edges that contain a certificate of its stored value, and (iii) each audit invalidates all active nodes reachable from the corrected roots and re-derives them in topological order using the current values of all structural parents $\mathrm{Pa}(v)$ required by $f_v$ (including parents omitted from the recorded certificate edges), recording fresh certificate edges. Then after every audit $t$, every active stored value satisfies $y_v(t^+)=F_v(X_{t^+})$, matching the task value induced by the currently active root assignment $X_{t^+}$ in memory (and thus equals the true task value once all wrong roots in $v$'s lineage have been audited). This holds for any audit order, including simultaneous corrections to multiple roots.
\end{proposition}

As long as no parent in $v$'s certificate is invalidated, those parents keep their formation-time values and force $f_v$ to retain its stored output regardless of changes to other parents (proof in Appendix~\ref{app:robustness}). For any unate rule, formation-time value-agreement parents form a valid certificate, whereas single-parent (\emph{but-for}) edges fail whenever support is overdetermined (e.g., three true inputs in an at-least-two-of-four rule yield zero but-for edges). Note that a \emph{certificate} governs which parent edges can be omitted at formation ($C_v\subsetneq\mathrm{Pa}(v)$), whereas \emph{update sensitivity} ($\Asens$) depends on the specific update and downstream propagation (Appendix~\ref{app:robustness}).

\section{Experimental Setup}
\label{sec:setup}

We evaluate these predictions in a controlled persistent-memory testbed (Table~\ref{tab:setup}) and on a public multi-hop memory benchmark (roadmap in Appendix Table~\ref{tab:roadmap}).

\subsection{Controlled persistent-memory lifecycle and depth sweep}
We generate $50$ frozen episodes paired across all conditions (Appendix~\ref{app:benchmark}). In each episode, a supply-chain assistant receives $24$ source facts (\emph{roots}) from six fictional sources, one per step ($t=0,\dots,23$), stating binary attributes of nonsense entity names so pretraining knowledge cannot be used. Exactly $18$ roots arrive with their true value and $6$ with an incorrect value; $20$ steps after arrival (\emph{audit delay} $d=20$), each root is audited and any incorrect root is replaced by its true value. As facts arrive, a reader LLM (\texttt{gpt-5.6-luna} or \texttt{deepseek-flash}, direct APIs) answers YES/NO questions over memory and writes each conclusion back. Six L1 questions each depend on four roots via unate rules ($(A\lor B)\land(C\lor D)$, $(A\land B)\lor(C\land D)$, at-least-two-of-four, at-least-three-of-four); L2 questions combine L1 pairs via binary unate rules ($x\land y$, $x\lor y$, $x\land\lnot y$, $\lnot x\land y$) retrieved with budget $K=2$ plus four distractors, and at $t=46$ all derived questions are re-asked as write-back-free probes.

To test scaling with reasoning depth $D\in\{1,\dots,5\}$, we obtain each depth-$D$ variant by truncating a shared depth-$5$ binary chain over the L1 conclusions after $D$ levels, padding with $5-D$ unrelated derived memories so total memory size stays fixed at ten derived items. We compare five formation-time provenance graphs under the same reachability repair operator: (1) \textbf{Oracle structural ($\Gtrue$)}, keeping all rule-required parents; (2) \textbf{LLM material-support}, prompting a frozen \texttt{deepseek-flash} attributor (or \texttt{gpt-5.6-luna} as a second attributor) with the stored conclusion, rule, and active candidates to select supporting memories; (3) \textbf{Value-agreement support}, keeping true parents whose signed value at formation matches $y_v$; (4) \textbf{Rule-based but-for}, keeping true parents whose single flip changes the rule output; and (5) \textbf{Matched random}, copying the LLM graph's exact per-node counts of true and false incoming edges uniformly at random. All conditions are evaluated in \textbf{replay mode} (on shared logged formations) and in \textbf{end-to-end closed-loop mode} (regenerating answers and refreshing edges live at every repair).

\begin{table}[t]
\caption{Summary of the primary controlled testbed.}
\label{tab:setup}
\centering
\small
\begin{tabular}{p{0.28\linewidth}p{0.65\linewidth}}
\toprule
Component & Configuration \\
\midrule
Episodes & 50 frozen episodes, paired across all depths and provenance conditions \\
Source facts (roots) & 24 per episode ($6\text{ sources}\times4$); 6 initially wrong; 8 binary attributes \\
Derived memories & 6 L1 nodes (4 parents each) + 3 base L2 nodes; depth sweep $D=1$--$5$ \\
Lifecycle schedule & 1 root arrives/step; audited at arrival $+20$; final probes at $t=46$ \\
Retrieval context & Rule-relevant parents ($K=2$ for L2) + 4 unrelated distractors per query \\
Reader / Attributor LLMs & Readers: \texttt{gpt-5.6-luna}, \texttt{deepseek-flash}; Attributors: \texttt{deepseek-flash}, \texttt{gpt-5.6-luna} \\
Repair \& Uncertainty & Reachability invalidation + topological re-derivation; 4,000-draw episode bootstrap 95\% CIs \\
\bottomrule
\end{tabular}
\vspace{-4pt}
\end{table}

\subsection{Public multi-hop memory benchmark}
To test whether selective LLM provenance generalizes beyond fixed-DAG Boolean rules to natural-language streams where updates dynamically rewire downstream lookups, we evaluate on MemoryAgentBench Conflict Resolution / FactConsolidation \citep{hu2026memoryagentbench}. Once a pre-update multi-hop path is complete, the agent answers in-stream and writes the intermediate hop entity to memory; after all updates finish, we probe the final question again. We compare four policies (Appendix~\ref{app:public}): \textbf{no write-back} (never storing intermediates), \textbf{keep-stale} (retaining intermediates without repair), \textbf{closure invalidation} (removing intermediates reachable from a corrected fact), and \textbf{invalidation + re-derivation} (recomputing and storing the invalidated intermediates). Our primary test uses a \emph{reversed-order control} where a counterfactual fact arrives first and the real-world fact arrives later as the correction, so pretraining knowledge favors the post-update answer and any penalty from a retained stale intermediate is conservative. Units are split by whether the update leaves the intermediate valid ($n=24$) or invalid ($n=38$).

\section{Results}
\label{sec:results}

\subsection{Outdated derived memories directly harm post-update answers}
\label{sec:res_ablation}
We first verify our motivating premise: leaving an outdated derived conclusion in memory degrades downstream answers even after its source fact is corrected. Across $462$ post-correction test queries (from the $49$ of $50$ episodes containing at least one post-correction stale conclusion; one episode had zero stale probes), correcting the source fact while keeping its stale derived conclusion in memory yields $0.907$ accuracy, whereas removing that stale conclusion raises accuracy to $0.996$ (paired gain $+0.089$, 95\% CI $[0.057, 0.123]$). Replacing the stale item with an unrelated derived statement of identical form and length keeps accuracy at $0.991$, while changing prompt order has only a minor effect ($0.868$ vs.\ $0.838$; Appendix Fig.~\ref{fig:app_stale_ablation}). As audit delay $d\in\{2,5,10,20\}$ grows, more conclusions form on top of the error, causing source-only updates to fall progressively further behind closure repair (Appendix Fig.~\ref{fig:app_delay}).

\subsection{As depth grows, structural coverage drops to half while stale coverage stays near 100\%}
\label{sec:res_depth}
Tracing LLM material-support provenance across reasoning depths $D=1,\dots,5$ (Fig.~\ref{fig:three_objects}B--C) shows that structural coverage ($C_{\mathrm{struct}}$) drops from $0.68$ at $D=1$ to $0.50$ at $D=5$ for Luna and from $0.68$ to $0.51$ for DeepSeek, omitting half the historical lineage at depth $5$. Yet this incomplete graph reaches nearly every memory that needs repair: update-sensitive coverage ($C_{\mathrm{sens}}$) stays at $0.95$--$0.99$ (Luna) and $1.00$ (DeepSeek), and stale-state coverage ($C_{\mathrm{stale}}$) stays at $0.95$--$1.00$ (Luna) and $1.00$ (DeepSeek). End-of-lifecycle outcomes track stale coverage: only $0$--$2.3\%$ of memories remain stale on Luna and $0\%$ on DeepSeek, and at $D=5$ final probe error is $0.10$ errors/episode under LLM provenance versus $0.04$ under oracle repair and $0.94$ without repair on Luna ($0.00$, $0.00$, and $0.86$ on DeepSeek; Table~\ref{tab:app_depth}).

\subsection{How reader errors separate update sensitivity from stored staleness}
\label{sec:three_disagree}
On the shared depth-sweep formations, DeepSeek makes zero formation errors, so $\Asens$ and $\Astale$ coincide on all $682$ active pairs, whereas structural descendants are nearly four times as numerous ($2{,}630$ pairs; Table~\ref{tab:app_three_set}). Low-effort Luna occasionally errs at formation, creating $35$ disagreements between $\Asens$ and $\Astale$ ($18$ sensitive-but-fresh, $17$ insensitive-but-stale). Because both the LLM attributor and value agreement record parents supporting the \emph{stored} answer $y_v$, when Luna mistakenly writes NO for a true YES node, they link the false parents supporting NO; if one of those parents is later corrected to true, repair reaches and fixes the node even though its ground-truth answer was YES all along (worked example in Appendix Table~\ref{tab:app_example}).

\subsection{Why half the lineage suffices: Certificates vs.\ random or counterfactual edges}
\label{sec:res_controls}
\begin{figure}[t]
  \centering
  \includegraphics[width=\linewidth]{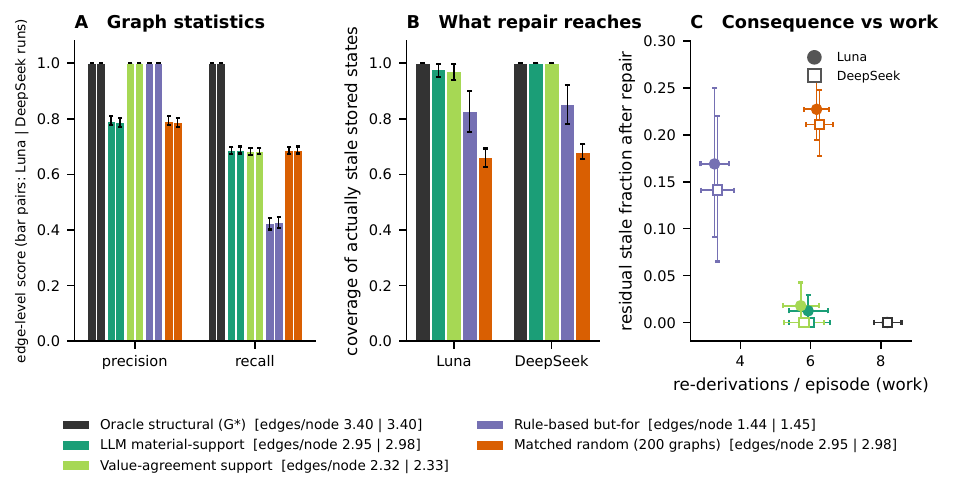}
  \caption{\textbf{Graphs with identical structural precision and recall can differ sharply in repair adequacy (replay controls).} \textbf{A:} Edge precision, recall, and budget across five provenance graphs. \textbf{B:} Stale-state coverage under fixed reachability repair: LLM $0.98/1.00$, value-agreement $0.97/1.00$, but-for $0.83/0.85$, matched random $0.66/0.68$ (Luna/DeepSeek). \textbf{C:} Residual stale fraction versus re-derivations per episode. Error bars are episode-cluster paired bootstrap 95\% CIs.}
  \label{fig:prov_compare}
  \vspace{-4pt}
\end{figure}

Comparing all five graphs on the shared depth-sweep formations shows that selective repair is not a byproduct of keeping roughly three edges per node (Fig.~\ref{fig:prov_compare}, Table~\ref{tab:app_provenance}). LLM provenance keeps $2.95$--$2.98$ edges per node ($0.79$ precision, $0.69$ recall, $5.9$--$6.0$ re-derivations/episode vs.\ $8.2$ for oracle replay). \textbf{Matched random} keeps the exact same number of true and false incoming edges at every node ($0.79$ precision, $0.69$ recall, $6.2$--$6.3$ re-derivations), yet its stale-state coverage plunges from $0.98/1.00$ to $0.66/0.68$ (Luna/DeepSeek), leaving $23\%/21\%$ residual stale memories versus $1.2\%/0\%$ for LLM provenance. The \textbf{rule-based but-for} graph has $1.00$ precision and $0.42$--$0.43$ recall, reaching only $0.83/0.85$ of stale state ($17\%/14\%$ residual stale) because $21\%$ of formations have overdetermined support where no single input flips the output at formation. By contrast, symbolic \textbf{value-agreement support} keeps all parents aligned with the stored conclusion ($1.00$ precision, $0.68$ recall), shares $69$--$72\%$ of edges with \texttt{deepseek-flash} ($72\%$ on the depth sweep), and reproduces the LLM repair profile: $0.971/1.00$ stale coverage, $1.8\%/0\%$ residual stale state, and $5.7/5.8$ re-derivations per episode.

\paragraph{A second LLM attributor shows that higher structural recall only adds work.}
Re-running attribution on all $2{,}389$ formations with \texttt{gpt-5.6-luna} raises structural precision/recall from $0.79/0.69$ to $0.82/0.78$ and structural coverage by $+0.12$--$0.13$ (Table~\ref{tab:app_attr}), but does not increase stale coverage ($0.965/1.000$ vs.\ $0.977/1.000$) while adding $+0.84$--$0.93$ re-derivations/episode. Both attributors retain $95$--$96\%$ of value-agreement (certificate) edges; \texttt{gpt-5.6-luna} simply keeps more true parents pointing \emph{against} the stored conclusion ($39$--$40\%$ vs.\ $11$--$12\%$), triggering extra re-derivations whose answers do not change.

\subsection{End-to-end closed-loop execution preserves the provenance ordering}
\label{sec:closed_loop}
\begin{figure}[t]
  \centering
  \includegraphics[width=\linewidth]{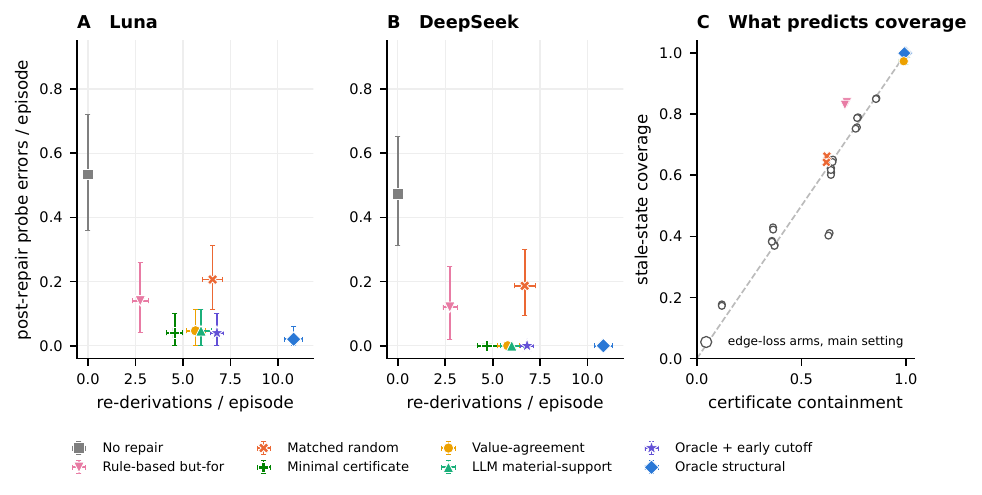}
  \caption{\textbf{End-to-end closed-loop runs preserve the provenance ordering} ($D\in\{1,3,5\}$ pooled, 50 episodes per reader). \textbf{A--B:} Post-repair probe error versus live re-derivations per episode for Luna (\textbf{A}) and DeepSeek (\textbf{B}). \textbf{C:} Across closed-loop and edge-loss conditions, stale-state coverage tracks certificate containment.}
  \label{fig:closed_loop}
  \vspace{-4pt}
\end{figure}

To verify that these findings hold when repairs actively update memory mid-stream, we ran every condition end to end in \textbf{closed loop} across $D\in\{1,3,5\}$ ($50$ episodes per reader): at each audit, invalidated nodes are re-answered live by the reader LLM and re-attributed before later audits occur (Fig.~\ref{fig:closed_loop}A--B, Appendix Table~\ref{tab:closed_loop}). Closed-loop oracle closure averages $10.8$ live re-derivations per episode with $0.02/0.00$ final probe errors (Luna/DeepSeek). Value-agreement ($5.7/5.8$ re-derivations, $0.05/0.00$ errors), LLM material-support ($6.0/6.0$ re-derivations, $0.05/0.00$ errors), and single \textbf{minimal-certificate} provenance ($4.6/4.7$ re-derivations, $0.04/0.00$ errors) all achieve near-oracle accuracy with $44$--$57\%$ fewer re-derivations. By contrast, closed-loop matched random reaches only $0.64/0.66$ of stale state ($0.21/0.19$ errors), and but-for leaves $0.14/0.12$ errors.

To connect these outcomes directly to Assumptions (i)--(ii) of Proposition~\ref{prop:certificate}, Table~\ref{tab:closed_loop} separates \textbf{local correctness} (whether the reader computed $y_v=f_v((y_p)_{p\in\mathrm{Pa}(v)})$ correctly from its stored parents) from \textbf{conditional certificate containment} (whether the recorded edges contain a certificate of $y_v$, given a locally correct formation; exact per-method formation denominators in Table~\ref{tab:failures}). On unate rules, both value-agreement and minimal-certificate provenance achieve $100\%$ conditional certificate containment on both readers ($98.9$--$99.0\%$ local correctness on Luna, $100\%$ on DeepSeek). LLM provenance (\texttt{deepseek-flash}) achieves $99.95\%$ conditional certificate containment on Luna and $99.81\%$ on DeepSeek, compared with $71.5\%/71.7\%$ for but-for and $62.5\%/62.2\%$ for matched random (Fig.~\ref{fig:closed_loop}C). Because LLM provenance omitted a certificate on $5$ locally correct formations ($1$ on Luna, $4$ on DeepSeek across $>4{,}100$ formations), Proposition~\ref{prop:certificate}'s worst-case guarantee does not apply unconditionally; however, in the observed correction sequences, those $5$ omissions produced $0$ additional missed stale states during repair (Appendix~\ref{app:failures}).

\paragraph{Repair work and total cost.}
Separate LLM attribution costs more tokens than it saves on these short Boolean tasks (Table~\ref{tab:cost}). Thinking-enabled \texttt{deepseek-flash} attribution uses $102\text{k}/101\text{k}$ tokens per episode, compared with $2{,}104/2{,}537$ saved repair tokens; low-effort \texttt{gpt-5.6-luna} uses $12.5\text{k}$ attribution tokens. Symbolic edge logging adds $1.8\,\text{ms}$ and $384\text{ bytes}$/episode.

\subsection{Public multi-hop memory: Stale intermediates harm accuracy when invalidated}
\label{sec:public}
\begin{figure}[t]
  \centering
  \includegraphics[width=0.64\linewidth]{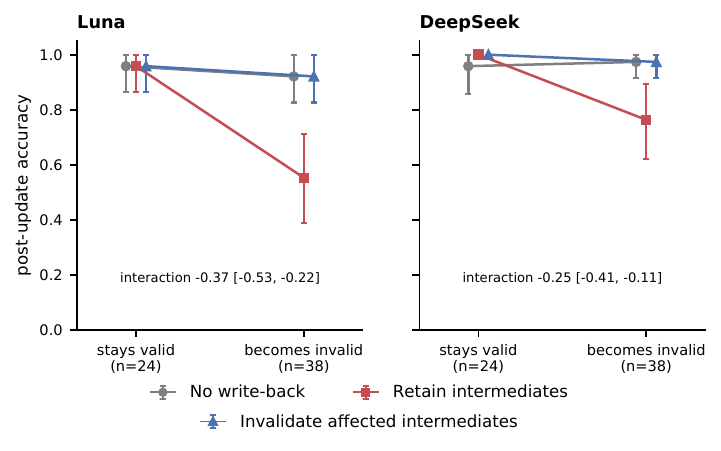}
  \caption{\textbf{On public multi-hop streams, persistent intermediates degrade accuracy specifically when invalidated by updates.} MemoryAgentBench FactConsolidation (reversed-order control), comparing no write-back, retention of intermediates, and closure invalidation across valid ($n=24$) and invalid ($n=38$) units. On invalid units, no write-back succeeds where retention fails in $14$ Luna and $8$ DeepSeek cases; the reverse never occurs. Error bars are paired bootstrap 95\% CIs.}
  \label{fig:public}
  \vspace{-4pt}
\end{figure}

Moving to natural-language multi-hop streams on MemoryAgentBench \citep{hu2026memoryagentbench}, we first test how stored intermediate hop conclusions affect post-update accuracy under oracle path provenance (Fig.~\ref{fig:public}). On the primary reversed-order control, when later updates leave the stored intermediate valid ($n=24$), \textbf{keep-stale} performs like \textbf{no write-back}, differing by $+0.00$ (Luna) and $+0.04$ (DeepSeek). When an update invalidates the intermediate ($n=38$), keeping the stale intermediate drops accuracy from $0.921$ (no write-back) to $0.553$ on Luna and from $0.974$ to $0.763$ on DeepSeek, whereas \textbf{closure invalidation} restores both models to the no-write-back baseline (write-back $\times$ validity interaction of $-0.37$ on Luna and $-0.25$ on DeepSeek; unit and component-clustered CIs in Appendix~\ref{app:public}, Table~\ref{tab:units_clustering}).

\subsection{LLM attribution achieves near-oracle accuracy on public data}
\label{sec:public_llm}
\begin{figure}[t]
  \centering
  \includegraphics[width=\linewidth]{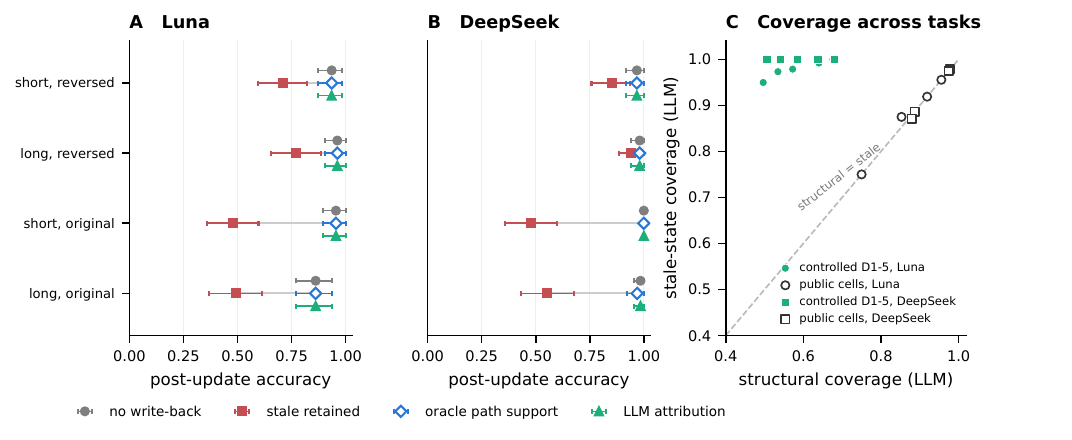}
  \caption{\textbf{LLM attribution on public multi-hop memory.} \textbf{A--B:} Accuracy across all evaluated units, including valid and invalid intermediates, for Luna (\textbf{A}) and DeepSeek (\textbf{B}). Short/long denote \texttt{mh\_6k}/\texttt{mh\_32k}, with $67/62$ and $65/52$ units in original/reversed order, respectively. \textbf{C:} Structural and stale-state coverage diverge under redundant logical rules and nearly coincide on the public lookup chains.}
  \label{fig:public_llm}
  \vspace{-4pt}
\end{figure}

Finally, we replace oracle path provenance on MemoryAgentBench with our frozen \texttt{deepseek-flash} attributor, which sees only the stored intermediate, the question, and the facts in memory at formation. Accuracy here includes all evaluated units in each split and order, both valid and invalid intermediates. Across all eight cells, LLM-attributed repair achieves accuracy within $0.00$--$0.02$ of oracle path repair (component-clustered 95\% CIs for the differences within $[0.00, 0.04]$), with $0.97$--$1.00$ edge precision, $0.88$--$0.96$ recall, and $0.42$--$0.56$ re-derivations/unit vs.\ $0.46$--$0.61$ for oracle support (Fig.~\ref{fig:public_llm}A--B, Table~\ref{tab:public_llm}).

MemoryAgentBench updates can rewire the downstream lookup path. This evaluation therefore tests generalization beyond Proposition~\ref{prop:certificate}'s fixed-DAG setting. Here, $94$--$100\%$ of structurally affected intermediates change ground-truth value after an upstream update. Structural and stale-state coverage nearly coincide ($0.75$--$0.98$ for both; identical in $6/8$ cells and within $0.01$--$0.03$ in the other two, Table~\ref{tab:public_llm}), contrasting with the redundant logical rules (Fig.~\ref{fig:public_llm}C). Despite incomplete stale coverage, aggregate accuracy equals oracle repair in seven cells and is higher by $0.015$ in one (95\% CI $[0.000,0.040]$).

\section{Discussion}
\label{sec:discussion}

\paragraph{Why certificates govern repair adequacy.}
Certificate-containing provenance is sufficient for repair under Proposition~\ref{prop:certificate}'s assumptions. When redundant support permits a proper-subset certificate, other structural edges can be omitted. Value agreement supplies a certificate on unate rules but fails on the evaluated non-unate rules ($0.41$--$0.42$ stale coverage; Appendix~\ref{app:sensitivity}). Minimal certificates and LLM attribution retain $0.99$--$1.00$ stale coverage there by keeping broader support, with smaller work savings than in the unate depth sweep. On the public lookup chains, structural and stale coverage nearly coincide.

\paragraph{Implications for system design and evaluation.}
The cost results distinguish repair work from total computation. Metrics that penalize unreached structural descendants (such as $\text{Leak\%}=1-C_{\mathrm{struct}}$) count both harmless omissions and harmful misses. Reporting stale-state coverage, residual stale state, repair work, total cost, and post-repair accuracy alongside structural fidelity makes these differences visible.

\section{Limitations}
\label{sec:limitations}

Root corrections arrive programmatically at known audit steps; autonomous error discovery is outside our scope. Proposition~\ref{prop:certificate} requires a fixed dependency DAG, local calculation correctness, certificate containment, and access to all required structural parents during re-derivation. On low-effort Luna, $21$ local calculation errors account for all $16$ unreached stale pairs; higher effort and \texttt{gpt-5.6-terra} each corrected $10/10$ unique failing prompts in diagnostic re-queries (Appendix~\ref{app:failures}). LLM attribution attains $99.81$--$99.95\%$ conditional certificate containment ($5$ omissions, $0$ resulting repair failures on the observed sequences). Omitting $25\%$ of required parents during retrieval lowers its stale coverage to $0.80$ (Appendix~\ref{app:sensitivity}). The large coverage gap under redundant rules narrows on public lookup chains with dynamically changing dependencies. Extending certificate tracking to free-form narrative summaries and multi-agent shared memories is a direction for future work.

\section{Conclusion}
\label{sec:conclusion}

When upstream facts change, the stored conclusions that become wrong can be much fewer than the structural descendants. In the controlled unate depth sweep, LLM and certificate provenance retain near-oracle accuracy with $44$--$57\%$ fewer closed-loop re-derivations. Non-unate and wide-graph results show that the savings depend on the available support redundancy and graph structure. On the public lookup chains, structural and stale-state coverage nearly coincide and LLM attribution achieves near-oracle accuracy with modest work savings. The useful distinction is between recovering historical dependencies and retaining enough support to repair the memories that are wrong.

\clearpage
\subsection*{Reproducibility statement}
The controlled testbed uses frozen paired episodes and deterministic lifecycle schedules. Appendix~\ref{app:benchmark} provides the notation table, experiment overview, episode generation procedure, rule families, provenance definitions, direct-API model configurations, and exact prompt templates. Appendices~\ref{app:controlled_full}--\ref{app:failures} provide complete numerical tables, mathematical proofs, machine verification counts, prior-method adaptations, sensitivity sweeps, token/latency cost accounting, and failure breakdowns.

\subsection*{AI use statement}
Generative AI tools were used across this project to brainstorm and refine research hypotheses, provide feedback on experimental design, assist with literature search, formulate mathematical claims and proofs, write and debug experiment and analysis code, format figures and tables, and draft and edit manuscript text. All experimental designs, code implementations, empirical results, mathematical proofs, citations, and manuscript text were verified by the authors, who take full responsibility for the contents of the paper.

\appendix
\setcounter{topnumber}{5}
\setcounter{bottomnumber}{5}
\setcounter{totalnumber}{8}
\renewcommand{\floatpagefraction}{0.8}
\raggedbottom

\section{Controlled Testbed, Notation, and Reproducibility Details}
\label{app:benchmark}

Tables~\ref{tab:notation} and~\ref{tab:roadmap} summarize the notation and experimental roadmap of the paper.

\begin{table}[!htbp]
\centering
\small
\caption{Summary of notation and key terms.}
\label{tab:notation}
\begin{tabular}{p{2.5cm}p{10.5cm}}
\toprule
Symbol / Term & Meaning \\
\midrule
$u$, $t$ & A corrected source fact (root) and its audit time; $X_{t^-}$ and $X_{t^+}$ denote the root value assignments immediately before and after correcting $u$ at time $t$. \\
$f_v$, $F_v$, $y_v$ & The deterministic local rule of derived node $v$ over its direct parents $\mathrm{Pa}(v)$, the induced root-to-node task function $F_v(X)$ over root assignment $X$, and the answer actually stored in memory for $v$ by the reader LLM. \\
$\Gtrue$, $\Gobs$ & The oracle structural graph (all rule-required parents presented at formation) and an observed provenance graph. \\
$\Astruct(u,t)$ & Active derived memories downstream of $u$ in $\Gtrue_t$ immediately before audit $t$ (Eq.~1). \\
$\Asens(u,t)$ & Structural descendants whose ground-truth value changes between $X_{t^-}$ and $X_{t^+}$ (Eq.~2). \\
$\Astale(u,t)$ & Structural descendants whose stored value $y_v(t^-)$ disagrees with the updated ground truth $F_v(X_{t^+})$ (Eq.~3). \\
$C_k$ & Coverage: fraction of $\mathcal{A}_k$ ($k\in\{\mathrm{struct},\mathrm{sens},\mathrm{stale}\}$) reached by repair over $\Gobs$, pooled across events (Eq.~4). \\
Residual stale & Fraction of derived memories formed under an unaudited wrong ancestor whose stored value disagrees with the final ground truth and is never re-derived. \\
Value agreement & Parents whose signed truth value matches the stored conclusion $y_v$. \\
Certificate & A parent subset whose values force the stored output under every completion of the remaining parents; subset-minimal certificates are sufficient reasons. \\
Replay vs.\ closed loop & \emph{Replay} evaluates graph reachability on a shared set of logged formations without regenerating reader answers; \emph{closed loop} (end-to-end) regenerates invalidated answers and refreshes attribution at every repair step. \\
\bottomrule
\end{tabular}
\end{table}

\begin{table}[!htbp]
\centering
\small
\caption{Overview of experimental evaluations and what each establishes.}
\label{tab:roadmap}
\begin{tabular}{p{4.1cm}p{8.9cm}}
\toprule
Experiment & Claim established \\
\midrule
Stale-content & Removing or replacing a stale derived memory restores post-update probe accuracy; delay increases error under source-only correction (Sec.~\ref{sec:res_ablation}, Figs.~\ref{fig:app_stale_ablation}--\ref{fig:app_delay}). \\
Depth sweep ($D=1$--$5$) & Structural coverage falls with depth while stale-state coverage and post-repair accuracy remain near oracle (Sec.~\ref{sec:res_depth}, Fig.~\ref{fig:three_objects}, Table~\ref{tab:app_depth}). \\
Graph-only controls (replay) & With formations held fixed, the identity of retained parents---not edge count or aggregate precision/recall---governs repair adequacy (Sec.~\ref{sec:res_controls}, Fig.~\ref{fig:prov_compare}). \\
End-to-end closed-loop runs & The provenance ordering holds when every repair regenerates reader answers and refreshes attribution (Sec.~\ref{sec:closed_loop}, Fig.~\ref{fig:closed_loop}, Table~\ref{tab:closed_loop}). \\
Second attributor comparison & Higher structural edge recall increases re-derivation work without improving stale-state coverage (Sec.~\ref{sec:res_controls}, Table~\ref{tab:app_attr}). \\
Propositions~\ref{prop:vc}--\ref{prop:certificate} \& simulation & Certificate support guarantees accuracy across sequential and simultaneous updates; value agreement is a certificate for unate rules (Sec.~\ref{sec:adequacy}, App.~\ref{app:robustness}, Table~\ref{tab:theorem_check}). \\
Non-unate and wide stress tests & Value agreement fails where it is not a certificate, while minimal certificates and LLM attribution remain adequate; separation grows with branching (App.~\ref{app:sensitivity}, Table~\ref{tab:stress}). \\
Public multi-hop benchmark & Stale intermediates harm accuracy only when invalidated by updates; LLM attribution matches oracle path support; lookup chains have no redundancy so $C_{\mathrm{struct}} \approx C_{\mathrm{stale}}$ (Secs.~\ref{sec:public}, \ref{sec:public_llm}, Figs.~\ref{fig:public}, \ref{fig:public_llm}). \\
Cost and failure accounting & Repair-work savings are separated from total attribution cost; higher effort and a stronger reader each correct the 10 selected failing prompts in diagnostic re-queries (Apps.~\ref{app:cost}--\ref{app:failures}). \\
\bottomrule
\end{tabular}
\end{table}

\paragraph{Episode generation.}
Episodes are procedurally generated from integer seeds (generator \texttt{gen\_v1}) and written to disk prior to any model call. Generation is purely programmatic: no language model is involved, and every natural-language string presented to the reader is populated via the fixed templates below. Entity names are synthesized by concatenating two nonsense syllables with a facility type suffix (e.g., \emph{Quendris Industries}, \emph{Pelldell Port}), ensuring that no entity or attribute value can be recalled from pretraining. We generated 200 seeds and use the first 50 (\texttt{ep0000}--\texttt{ep0049}) across all reported controlled experiments.

\paragraph{Source claims and audit schedule.}
Each episode contains 24 roots from six fictional sources (four claims per source). The eight binary attribute types are: company revenue trend (up/down), company margin trend (up/down), supplier delivery status (on\_time/delayed), port operating status (open/disrupted), factory production status (normal/reduced), warehouse inventory level (adequate/low), carrier capacity (available/constrained), and regulator permit status (approved/pending). Each root's true value is drawn uniformly, and exactly six roots per episode are initialized with the opposite of their ground-truth value. The LOW placement assigns one incorrect root to each of the six sources; the HIGH placement assigns three incorrect roots to each of two sources. Claims arrive one per step from $t=0$ to $t=23$. Every root is audited 20 steps after arrival, at which point an incorrect root is replaced by its ground-truth value. Post-audit probes are issued at $t=46$, after the final audit.

\paragraph{Derived-state rules.}
The six L1 nodes partition the 24 roots at random (four parents per node). Each parent enters as a literal testing whether the root takes a sampled trigger value (e.g., ``the port is \emph{open}''). Rule families are $(A\lor B)\land(C\lor D)$, $(A\land B)\lor(C\land D)$, at-least-two-of-four, and at-least-three-of-four, sampled uniformly (82/75/71/72 of the 300 L1 nodes across the 50 episodes). The three base L2 nodes each combine two L1 conclusions using one of $x\land y$, $x\lor y$, $x\land\lnot y$, $\lnot x\land y$. Generation rejection-samples until exactly three of six L1 nodes evaluate to true and the three L2 nodes are not all identical under the true roots. An L1 query is issued one step after its last parent arrives plus a random lag of 0--2 steps; an L2 query is issued after both L1 parents in the same manner. Each episode also includes 12 background facts about unrelated entities drawn from four disjoint attribute types (headcount trend, audit rating, credit outlook, fleet status); every query context includes four of them as distractors that never support any node.

\paragraph{Oracle structural graph and lifecycle labels.}
For a materialized node $v$, $\Gtrue$ connects $v$ to the rule-required memories shown at formation. Structural descendants are computed transitively over active derived memories. At each incorrect-root audit, update sensitivity is evaluated using the root assignment immediately before and after that correction, and staleness compares the active stored reader value immediately before audit against the post-audit ground-truth task value. States already replaced by earlier repairs are excluded.

\paragraph{Depth construction.}
The depth sweep constructs a nested chain from the six L1 nodes. The first leaf $L1_a$ is the L1 node containing the incorrect root audited last, ensuring that an uncorrected error is active when the chain forms; the next four L1 nodes by index serve as the remaining leaves. The chain is defined by $C_2=\mathrm{op}_2(L1_a,L1_b)$ and $C_k=\mathrm{op}_k(C_{k-1},L1_k)$ for $k=3,\dots,5$, with binary operations drawn from the four L2 connectives. Chain operations are sampled once per base episode and shared across depth variants, so the depth-$D$ graph is a prefix of the depth-5 graph ($D=1$ queries $L1_a$ alone). All chain nodes form at the same step (one step after the final L1 query), and $5-D$ unrelated padding entries keep the number of derived memories fixed at ten across depths.

\paragraph{Worked example.}
Table~\ref{tab:app_example} illustrates node L1\_5 of \texttt{ep0017} under LOW placement in the base setting. Its question asks whether Project Volans is \emph{launch ready} under the rule at-least-two-of-four. The node is queried at $t=11$, one step after its last parent r01 arrives with the incorrect claim \emph{down}. Because literals $C$ and $D$ hold, the ground-truth task value at formation is YES.

\begin{table}[!htbp]
\caption{Parents of node L1\_5 (\texttt{ep0017}, LOW). Only r01 is an incorrect claim.}
\label{tab:app_example}
\centering
\scriptsize
\begin{tabular}{lllllllr}
\toprule
Literal & Root & Entity & Attribute & Trigger & True & Claimed & Arrival / correction \\
\midrule
$A$ & r01 & Rheltane Industries & margin trend & up & up & \textbf{down} & 10 / 30 \\
$B$ & r03 & Ulmdell Port & operating status & disrupted & open & open & 1 / -- \\
$C$ & r14 & Rheloth Freight & capacity & constrained & constrained & constrained & 9 / -- \\
$D$ & r17 & Galdoth Industries & margin trend & down & down & down & 3 / -- \\
\bottomrule
\end{tabular}
\end{table}

The Luna reader erroneously answered and stored NO. Given this stored conclusion, the material-support attributor returned r01 and r03, the two false literals supporting NO; the symbolic value-agreement graph retains the exact same two edges. Auditing r01 at $t=30$ flips literal $A$ to true, so three literals hold instead of two; the ground-truth value remains YES and L1\_5 $\notin\Asens(\text{r01},30)$. However, because the stored answer NO disagrees with the post-audit truth YES, L1\_5 $\in\Astale(\text{r01},30)$: an insensitive-but-stale state caused by a reader error. Because both the LLM attributor and value agreement conditioned on the stored answer NO and linked r01 to L1\_5, the audit at $t=30$ reached and re-derived L1\_5 to YES, correcting the downstream probe at $t=46$. A graph constructed from the ground-truth output YES would have linked r14 and r17 instead and missed this repair.

\paragraph{Model configurations.}
All model calls in this paper use the direct OpenAI and DeepSeek APIs. The controlled readers are \texttt{gpt-5.6-luna} (low reasoning effort, strict JSON output) and \texttt{deepseek-flash} (temperature 0, JSON output); both allow 4,000 output tokens with one 32,000-token retry. The primary material-support attributor is \texttt{deepseek-flash} with thinking enabled, JSON output, a 4,000-token limit, and one 32,000-token retry. The second attributor is \texttt{gpt-5.6-luna} called via the direct API with low reasoning effort and identical prompt bytes across all 2,389 formation prompts (0 unresolved).

\paragraph{Statistical procedures.}
Controlled confidence intervals use paired episode-cluster bootstrap resampling with 4,000 draws; all series compared within a reader/depth cell share the same resampled episode indices. Public benchmark intervals use both unit-cluster and connected-component-cluster paired bootstrap resampling (Table~\ref{tab:units_clustering}).

\paragraph{Prompt templates.}
The controlled reader receives the following system prompt:
\begin{small}
\begin{verbatim}
You are the reasoning module of a supply-chain monitoring assistant
that keeps a persistent memory. Answer the question using the decision
rule and the memory entries provided. Memory entries are the
assistant's current stored beliefs; they may be incomplete. If a needed
fact is missing, make your best judgment from what is available.
Return JSON with fields: answer ("YES" or "NO"), confidence (0.0-1.0),
used_memory_ids (list of memory ids you relied on).
\end{verbatim}
\end{small}
The user message provides the current step, the question ``is \{name\}(\{subject\}) YES or NO?'', the decision rule, and the retrieved memory entries ordered by a deterministic hash of episode, node, and memory ID. The material-support attributor receives the following system prompt:
\begin{small}
\begin{verbatim}
You audit the provenance of a derived conclusion stored in an
assistant's memory. You are given the conclusion, the question and
decision rule it answered, and the candidate memory items that existed
when it was recorded. Identify which candidate items materially support
the conclusion (items whose content the conclusion depends on). Return
JSON: {"supports": [{"id": "<candidate id>", "confidence": <0-1>}]}.
attr_v1
\end{verbatim}
\end{small}
The attributor user message provides the recorded conclusion, its formation time, the question and decision rule, and all active memory items at formation time except the conclusion itself, sorted by ID. Returned IDs are intersected with valid candidate IDs.

\FloatBarrier
\section{Full Controlled Results}
\label{app:controlled_full}

\begin{figure}[H]
  \centering
  \includegraphics[width=\linewidth]{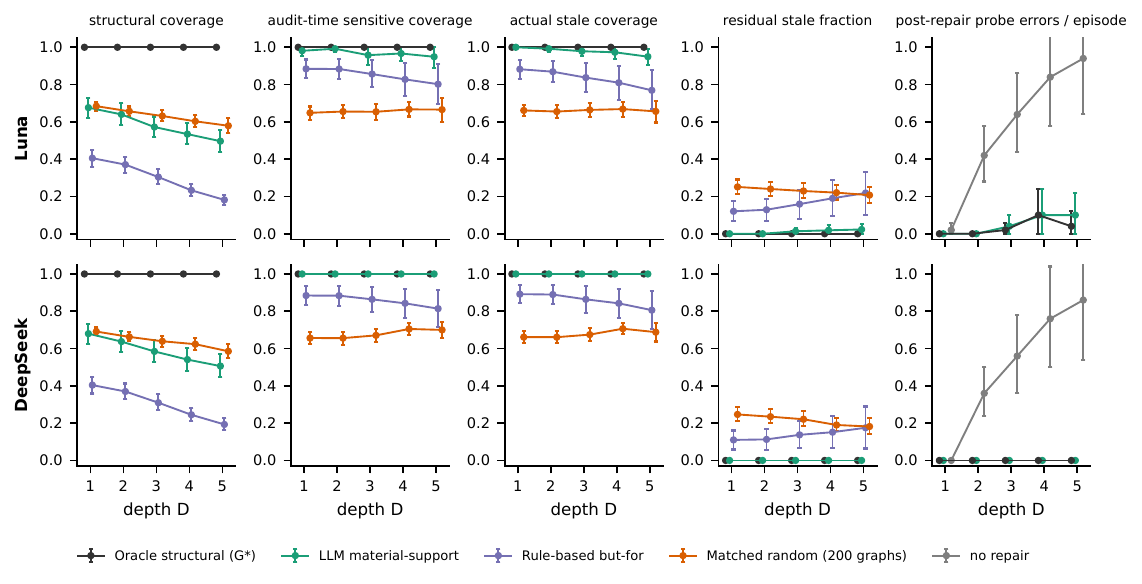}
  \caption{\textbf{Full reader-by-depth results.} Structural, update-sensitive, and stale-state coverage, residual stale fraction, and post-repair probe errors per episode across $D=1,\dots,5$. Error bars show episode-cluster paired bootstrap 95\% CIs over 50 episodes per reader and depth.}
  \label{fig:app_depth_full}
\end{figure}

\begin{figure}[H]
  \centering
  \includegraphics[width=0.86\linewidth]{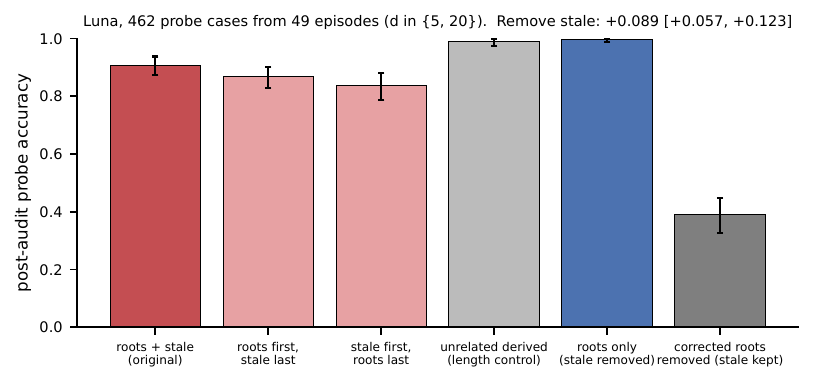}
  \caption{\textbf{Stale derived memories are causally harmful in post-audit probes (direct API).} Removing the stale derived item raises accuracy from $0.907$ to $0.996$ (paired gain $+0.089$ $[0.057, 0.123]$ across 462 probe cases from the 49 of 50 episodes with at least one post-correction stale conclusion); replacing it with an unrelated derived item of identical format yields $0.991$, whereas prompt-order changes give $0.868$ vs.\ $0.838$. Error bars are episode-cluster paired bootstrap 95\% CIs.}
  \label{fig:app_stale_ablation}
\end{figure}

\begin{figure}[H]
  \centering
  \includegraphics[width=0.88\linewidth]{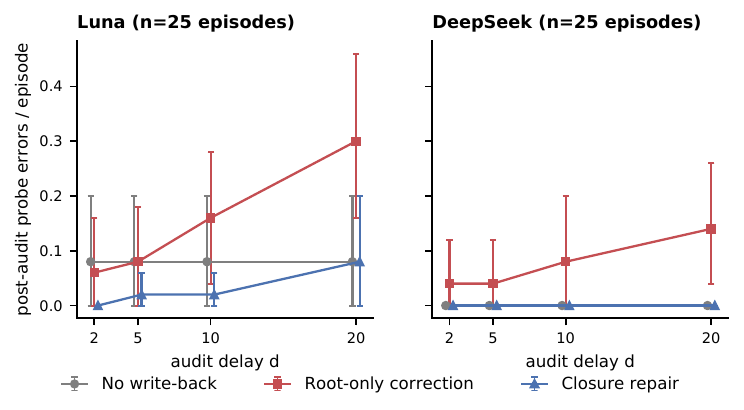}
  \caption{\textbf{Source-only correction diverges from provenance-closure repair as audit delay grows.} Post-audit probe errors per episode for no write-back, root-only correction that retains derived memories, and provenance-closure repair on the 25 episodes evaluated by both readers. At delay $d=20$, root-only minus closure error is $0.22$ $[0.04, 0.40]$ for Luna and $0.14$ $[0.04, 0.26]$ for DeepSeek.}
  \label{fig:app_delay}
\end{figure}

\begin{figure}[H]
  \centering
  \includegraphics[width=\linewidth]{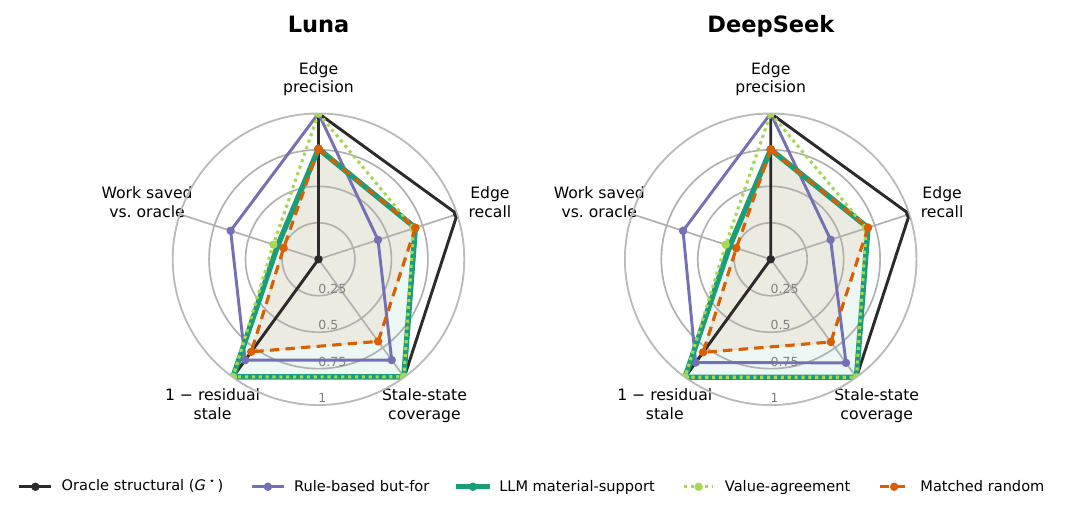}
  \caption{\textbf{Provenance profiles in the base $K=2$ setting} (Table~\ref{tab:app_provenance}, second block). LLM material-support and matched random coincide on edge precision and recall by construction, yet separate sharply on stale-state coverage and residual stale state; value-agreement support traces the LLM profile with perfect edge precision.}
  \label{fig:app_radar}
\end{figure}

Figure~\ref{fig:app_depth_full} and Table~\ref{tab:app_depth} report the complete reader-by-depth sweep. Figures~\ref{fig:app_stale_ablation}--\ref{fig:app_radar} and Tables~\ref{tab:app_provenance}--\ref{tab:app_attr} present the direct-API stale-content ablation, audit-delay sweep, replay provenance controls, and second-attributor comparison.

\begin{table}[H]
\caption{Reader-specific LLM-provenance depth sweep. Brackets indicate episode-cluster bootstrap 95\% CIs.}
\label{tab:app_depth}
\centering
\scriptsize
\resizebox{\linewidth}{!}{\begin{tabular}{llccccccc}
\toprule
Reader & $D$ & Struct.\ cov. & Sens.\ cov. & Stale cov. & Residual stale & \multicolumn{3}{c}{Probe errors / episode} \\
\cmidrule(lr){7-9}
 & & (LLM) & (LLM) & (LLM) & (LLM) & no repair & LLM & oracle \\
\midrule
Luna & 1 & 0.68 [0.62, 0.73] & 0.98 [0.95, 1.00] & 1.00 [1.00, 1.00] & 0.00 [0.00, 0.00] & 0.02 & 0.00 & 0.00 \\
 & 2 & 0.64 [0.58, 0.70] & 0.99 [0.97, 1.00] & 0.99 [0.97, 1.00] & 0.00 [0.00, 0.00] & 0.42 & 0.00 & 0.00 \\
 & 3 & 0.57 [0.52, 0.63] & 0.96 [0.91, 1.00] & 0.98 [0.96, 1.00] & 0.01 [0.00, 0.03] & 0.64 & 0.04 & 0.02 \\
 & 4 & 0.53 [0.48, 0.59] & 0.97 [0.93, 1.00] & 0.97 [0.94, 1.00] & 0.02 [0.00, 0.05] & 0.84 & 0.10 & 0.10 \\
 & 5 & 0.50 [0.44, 0.56] & 0.95 [0.89, 1.00] & 0.95 [0.90, 0.99] & 0.02 [0.00, 0.05] & 0.94 & 0.10 & 0.04 \\
\midrule
DeepSeek & 1 & 0.68 [0.62, 0.73] & 1.00 [1.00, 1.00] & 1.00 [1.00, 1.00] & 0.00 [0.00, 0.00] & 0.00 & 0.00 & 0.00 \\
 & 2 & 0.64 [0.58, 0.70] & 1.00 [1.00, 1.00] & 1.00 [1.00, 1.00] & 0.00 [0.00, 0.00] & 0.36 & 0.00 & 0.00 \\
 & 3 & 0.58 [0.53, 0.64] & 1.00 [1.00, 1.00] & 1.00 [1.00, 1.00] & 0.00 [0.00, 0.00] & 0.56 & 0.00 & 0.00 \\
 & 4 & 0.54 [0.48, 0.60] & 1.00 [1.00, 1.00] & 1.00 [1.00, 1.00] & 0.00 [0.00, 0.00] & 0.76 & 0.00 & 0.00 \\
 & 5 & 0.51 [0.45, 0.57] & 1.00 [1.00, 1.00] & 1.00 [1.00, 1.00] & 0.00 [0.00, 0.00] & 0.86 & 0.00 & 0.00 \\
\bottomrule
\end{tabular}}
\end{table}

\begin{table}[H]
\caption{Replay provenance graph statistics and repair outcomes on shared formations from the LLM run. The first block pools the depth sweep $D=1$--$5$ within episode; the second reports the base $K=2$ setting. Note that the oracle re-derivation count here ($8.2$) replays the oracle graph on the LLM run's formations; closed-loop oracle execution averages $10.8$ re-derivations per episode (Table~\ref{tab:closed_loop}).}
\label{tab:app_provenance}
\centering
\scriptsize
\resizebox{\linewidth}{!}{\begin{tabular}{lllcccccc}
\toprule
Setting & Reader & Provenance & Edges/node & Edge P & Edge R & Stale cov. & Residual stale & Re-deriv./ep. \\
\midrule
depth sweep D1-5 & Luna & Oracle structural ($G^\star$) & 3.40 & 1.00 & 1.00 & 1.00 [1.00, 1.00] & 0.00 [0.00, 0.00] & 8.2 \\
 &  & LLM material-support & 2.95 & 0.79 & 0.69 & 0.98 [0.95, 1.00] & 0.01 [0.00, 0.03] & 5.9 \\
 &  & Value-agreement support & 2.32 & 1.00 & 0.68 & 0.97 [0.94, 1.00] & 0.02 [0.00, 0.04] & 5.7 \\
 &  & Rule-based but-for & 1.44 & 1.00 & 0.42 & 0.83 [0.75, 0.90] & 0.17 [0.09, 0.25] & 3.3 \\
 &  & Matched random & 2.95 & 0.79 & 0.69 & 0.66 [0.63, 0.69] & 0.23 [0.19, 0.26] & 6.2 \\
\midrule
 & DeepSeek & Oracle structural ($G^\star$) & 3.40 & 1.00 & 1.00 & 1.00 [1.00, 1.00] & 0.00 [0.00, 0.00] & 8.2 \\
 &  & LLM material-support & 2.98 & 0.79 & 0.69 & 1.00 [1.00, 1.00] & 0.00 [0.00, 0.00] & 6.0 \\
 &  & Value-agreement support & 2.33 & 1.00 & 0.68 & 1.00 [1.00, 1.00] & 0.00 [0.00, 0.00] & 5.8 \\
 &  & Rule-based but-for & 1.45 & 1.00 & 0.43 & 0.85 [0.78, 0.92] & 0.14 [0.06, 0.22] & 3.3 \\
 &  & Matched random & 2.98 & 0.79 & 0.69 & 0.68 [0.65, 0.71] & 0.21 [0.18, 0.25] & 6.3 \\
\midrule
main setting & Luna & Oracle structural ($G^\star$) & 3.29 & 1.00 & 1.00 & 1.00 [1.00, 1.00] & 0.00 [0.00, 0.00] & 8.6 \\
 &  & LLM material-support & 3.04 & 0.76 & 0.70 & 1.00 [0.99, 1.00] & 0.00 [0.00, 0.01] & 6.1 \\
 &  & Value-agreement support & 2.25 & 1.00 & 0.68 & 1.00 [0.99, 1.00] & 0.00 [0.00, 0.01] & 5.8 \\
 &  & Rule-based but-for & 1.41 & 1.00 & 0.43 & 0.85 [0.80, 0.90] & 0.15 [0.10, 0.19] & 3.2 \\
 &  & Matched random & 3.04 & 0.76 & 0.70 & 0.70 [0.67, 0.72] & 0.22 [0.19, 0.24] & 6.5 \\
\midrule
 & DeepSeek & Oracle structural ($G^\star$) & 3.29 & 1.00 & 1.00 & 1.00 [1.00, 1.00] & 0.00 [0.00, 0.00] & 8.6 \\
 &  & LLM material-support & 3.07 & 0.75 & 0.70 & 1.00 [1.00, 1.00] & 0.00 [0.00, 0.00] & 6.1 \\
 &  & Value-agreement support & 2.26 & 1.00 & 0.69 & 1.00 [1.00, 1.00] & 0.00 [0.00, 0.00] & 5.8 \\
 &  & Rule-based but-for & 1.42 & 1.00 & 0.43 & 0.88 [0.84, 0.92] & 0.12 [0.08, 0.17] & 3.2 \\
 &  & Matched random & 3.07 & 0.75 & 0.70 & 0.70 [0.67, 0.73] & 0.21 [0.19, 0.24] & 6.5 \\
\bottomrule
\end{tabular}}
\end{table}

\begin{table}[H]
\caption{Second attributor comparison on identical formations (depth sweep $D=1$--$5$ pooled, direct API). Rows are grouped by the reader whose runs supplied the formations; differences are \texttt{gpt-5.6-luna} minus \texttt{deepseek-flash} with episode-cluster paired bootstrap 95\% CIs over 50 episodes. On the depth sweep, the two attributors share $72$--$73\%$ of edges (Jaccard), \texttt{deepseek-flash} shares $72\%$ with value-agreement support ($69$--$72\%$ when pooling depth and base settings), and \texttt{gpt-5.6-luna} shares $67\%$.}
\label{tab:app_attr}
\centering
\scriptsize
\begin{tabular}{llccc}
\toprule
Reader runs & Metric & DeepSeek attributor & Luna attributor & Difference [95\% CI] \\
\midrule
Luna & Edge precision & 0.79 & 0.82 & $+0.03$ [0.01, 0.05] \\
 & Edge recall & 0.69 & 0.78 & $+0.09$ [0.08, 0.10] \\
 & Structural coverage & 0.56 & 0.68 & $+0.12$ [0.09, 0.16] \\
 & Sensitive coverage & 0.968 & 0.971 & $+0.003$ [$-0.002$, 0.012] \\
 & Stale coverage & 0.977 & 0.965 & $-0.012$ [$-0.030$, $-0.000$] \\
 & Residual stale & 0.012 & 0.025 & $+0.012$ [0.000, 0.030] \\
 & Re-derivations / episode & 5.93 & 6.78 & $+0.84$ [0.55, 1.17] \\
\midrule
DeepSeek & Edge precision & 0.79 & 0.82 & $+0.04$ [0.02, 0.06] \\
 & Edge recall & 0.69 & 0.78 & $+0.09$ [0.08, 0.11] \\
 & Structural coverage & 0.57 & 0.70 & $+0.13$ [0.09, 0.17] \\
 & Sensitive / stale coverage & 1.00 & 1.00 & 0.000 [0.000, 0.000] \\
 & Residual stale & 0.000 & 0.000 & 0.000 [0.000, 0.000] \\
 & Re-derivations / episode & 5.97 & 6.90 & $+0.93$ [0.63, 1.26] \\
\midrule
\multicolumn{5}{p{0.92\linewidth}}{Recall of value-agreement edges: 0.95 (DeepSeek attributor), 0.96 (Luna attributor). Recall of true edges pointing against the stored conclusion: 0.11--0.12 (DeepSeek attributor), 0.39--0.40 (Luna attributor).}\\
\bottomrule
\end{tabular}
\end{table}

\FloatBarrier
\section{Proofs, Certificate Theory, and Robustness Checks}
\label{app:robustness}

\paragraph{Proof of Proposition~\ref{prop:vc}.}
We induct over a topological order of $u$'s structural descendants. For any node $w$, write $F_w(X)$ for its ground-truth value under root assignment $X$ (with $F_r(X)=X_r$ for roots), so $F_v(X)=f_v((F_p(X))_{p\in\mathrm{Pa}(v)})$. Let $v\in\Asens(u,t)$ and write $x,x'$ for the signed parent vectors of $v$ under $X_{t^-}$ and $X_{t^+}$, normalized so that the local rule $f_v$ is nondecreasing in each coordinate. Suppose $F_v(X_{t^-})=f_v(x)=1$; the case $f_v(x)=0$ is symmetric. If every parent $p\in\mathrm{Pa}(v)$ that changed between $X_{t^-}$ and $X_{t^+}$ had $x_p=0$, all changes would be from $0$ to $1$, yielding $x'\ge x$ coordinatewise and by monotonicity $f_v(x')\ge f_v(x)=1$, which contradicts $F_v(X_{t^+})=f_v(x')\neq f_v(x)=F_v(X_{t^-})$. Hence at least one changed parent $p\in\mathrm{Pa}(v)$ satisfies $x_p=1=f_v(x)$, meaning $(p\rightarrow v)$ is a value-agreement edge in $G^{\mathrm{va}}_{X_{t^-}}$. Because $u$ is the only root that changes at audit $t$, $p$ is either $u$ itself or a derived descendant whose task value changes ($p\in\Asens(u,t)$), which is reachable from $u$ in $G^{\mathrm{va}}_{X_{t^-}}$ by the induction hypothesis. Since stored values equal task values under $X_{t^-}$, value agreement on stored values and on task values coincide. $\square$

\paragraph{Connection to minimal witnesses and sufficient reasons.}
Write a unate local rule as a monotone Boolean function $g(z)$ of signed parent values, with output $b=g(z)$. A sufficient reason is a subset-minimal set $W$ of coordinates such that fixing $z_W$ forces output $b$ under every completion of the remaining coordinates \citep{ignatiev2019abduction,darwiche2020reasons}; for positive outputs of monotone Boolean queries, these coincide with minimal witnesses \citep{buneman2001why}. Let $V_b(z)=\{i:z_i=b\}$ be the value-agreement set and $\mathcal{W}_b(z)$ the family of sufficient reasons. Then
\[
\bigcup_{W\in\mathcal{W}_b(z)} W \;\subseteq\; V_b(z).
\]
For $b=1$, suppose a candidate reason fixes a coordinate to $0$; removing that coordinate preserves sufficiency because setting it to $0$ already yields $1$ and raising it to $1$ cannot lower a monotone function. The case $b=0$ is symmetric. The inclusion can be strict: for $g(z)=z_1\lor(z_2\land z_3)$ at $z=(1,1,0)$, the unique minimal sufficient reason is $\{1\}$, whereas $V_1(z)=\{1,2\}$.

\paragraph{Value agreement is a certificate for unate rules.}
Normalize signs so that the local rule $f_v$ is nondecreasing, let $b=f_v(x)$, and let $V=\{i:x_i=b\}$. For $b=1$, any completion $z$ of the coordinates outside $V$ alters only coordinates where $x_i=0$, so $z\ge x$ coordinatewise and monotonicity forces $f_v(z)\ge f_v(x)=1$. The case $b=0$ is symmetric. Thus $V$ is a certificate of $f_v(x)$. We verified this property exhaustively across all $168$ monotone Boolean functions on four inputs and all $16$ input assignments ($2{,}688$ cases, $0$ violations). Conversely, value agreement is not a certificate for non-unate rules: it fails on $15/16$ assignments of four-input parity, $11/16$ for exactly-two-of-four, and $2/16$ for $(A\land B)\lor(\lnot A\land C)\lor D$.

\paragraph{Proof of Proposition~\ref{prop:certificate}.}
Fix the structural DAG and deterministic local rules $f_v$ over the correction lifecycle (where audits modify only root values and derived nodes change only via formation and re-derivation). We induct over the sequence of audits and, within each audit, over a topological order of the DAG. At any point with active root assignment $X$, the induction invariant is that every active derived node $v$ stores $y_v = f_v((y_p)_{p\in\mathrm{Pa}(v)}) = F_v(X)$ and has a subset of incoming recorded edges $C_v\subseteq\mathrm{Pa}(v)$ that forms a certificate for $f_v$ at $(y_p)_{p\in\mathrm{Pa}(v)}$, where every parent $p\in C_v$ is currently active and holds the same stored value $y_p$ as when $v$ was formed. This invariant holds at initial formation by assumptions (i) and (ii). Now consider an audit $t$ that updates a set of roots $U$ (changing the active root assignment from $X_{t^-}$ to $X_{t^+}$), and let $R$ denote the set of active nodes reachable from $U$ in the recorded graph $\Gobs$. We verify the invariant after audit $t$ by induction over a topological order of the DAG. If an active node $v\notin R$, then its certificate $C_v$ intersects neither $U$ nor $R$ (otherwise $v$ would be reachable from $U$ in $\Gobs$). Consequently, every parent $p\in C_v$ remains active and un-rederived, retaining the exact stored value $y_p(t^+)=y_p(t^-)$ it held when $v$ was formed. By the topological induction hypothesis for $p\notin R$, $y_p(t^+)=F_p(X_{t^+})$. Because fixing the parents in $C_v$ to those values forces $f_v$ to equal $y_v(t^-)$ under any completion of the remaining parents in $\mathrm{Pa}(v)\setminus C_v$, we have $F_v(X_{t^+}) = f_v((F_p(X_{t^+}))_{p\in\mathrm{Pa}(v)}) = y_v(t^-) = y_v(t^+)$, and $C_v$ remains a valid active certificate. If $v\in R$, every structural parent $p\in\mathrm{Pa}(v)$ precedes $v$ in topological order and therefore already satisfies $y_p(t^+)=F_p(X_{t^+})$ by the topological induction hypothesis; because re-derivation has access to the current values of all structural parents $\mathrm{Pa}(v)$ required by $f_v$ (including parents omitted from the prior certificate $C_v$), re-deriving $v$ from its current parent values produces $y_v(t^+)=f_v((y_p(t^+))_{p\in\mathrm{Pa}(v)})=F_v(X_{t^+})$ and records a fresh certificate of active memories. Note that $F_v(X_{t^+})$ is the task value under the currently active root assignment $X_{t^+}$ in memory; once all initially incorrect roots in $v$'s structural lineage have been audited, $X_{t^+}$ agrees with the ground-truth root assignment on $v$'s ancestors and $y_v(t^+)$ equals the final ground-truth task value. $\square$

\paragraph{Machine verification across rule families.}
To verify Proposition~\ref{prop:certificate} empirically, we simulated full lifecycles on randomly generated acyclic systems with six roots and six derived nodes ($20{,}000$ random systems per rule family), applying corrections either one root at a time (single) or two roots simultaneously (paired). Table~\ref{tab:theorem_check} shows that certificate edges leave zero systems with stale state across unate rules, non-unate rules, arbitrary Boolean truth tables, and three-valued numeric rules (maximum and sum-threshold), while reducing re-derivations relative to oracle closure. By contrast, but-for and random edges leave thousands of systems with stale state in every family, and value-agreement edges leave $0/20{,}000$ stale systems on unate rules but fail as predicted on non-unate and general table rules.

\begin{table}[!htbp]
\caption{Lifecycle simulation check of Proposition~\ref{prop:certificate} across $20{,}000$ random systems per family under single-root and paired simultaneous corrections.}
\label{tab:theorem_check}
\centering
\scriptsize
\begin{tabular}{llcccc}
\toprule
Rule family & Edges & Systems with stale (single) & Systems with stale (paired) & Re-deriv./system (single) & (paired) \\
\midrule
unate & oracle & 0 / 20,000 & 0 / 20,000 & 11.30 & 8.30 \\
 & cert & 0 / 20,000 & 0 / 20,000 & 4.57 & 4.14 \\
 & agree & 0 / 20,000 & 0 / 20,000 & 6.31 & 5.47 \\
 & butfor & 6,563 / 20,000 & 6,092 / 20,000 & 2.23 & 2.08 \\
 & random & 8,675 / 20,000 & 6,749 / 20,000 & 7.19 & 6.00 \\
\midrule
nonunate & oracle & 0 / 20,000 & 0 / 20,000 & 11.25 & 8.28 \\
 & cert & 0 / 20,000 & 0 / 20,000 & 6.63 & 5.52 \\
 & agree & 13,232 / 20,000 & 11,079 / 20,000 & 5.43 & 4.79 \\
 & butfor & 5,247 / 20,000 & 4,737 / 20,000 & 4.75 & 3.94 \\
 & random & 12,674 / 20,000 & 9,968 / 20,000 & 7.16 & 6.01 \\
\midrule
table & oracle & 0 / 20,000 & 0 / 20,000 & 11.34 & 8.35 \\
 & cert & 0 / 20,000 & 0 / 20,000 & 6.57 & 5.56 \\
 & agree & 16,508 / 20,000 & 14,825 / 20,000 & 4.42 & 3.98 \\
 & butfor & 6,642 / 20,000 & 6,000 / 20,000 & 4.72 & 4.25 \\
 & random & 12,738 / 20,000 & 10,032 / 20,000 & 7.15 & 6.00 \\
\midrule
numeric & oracle & 0 / 20,000 & 0 / 20,000 & 11.26 & 8.31 \\
 & cert & 0 / 20,000 & 0 / 20,000 & 5.42 & 4.70 \\
 & butfor & 3,973 / 20,000 & 3,720 / 20,000 & 3.29 & 2.94 \\
 & random & 8,278 / 20,000 & 6,513 / 20,000 & 7.18 & 6.03 \\
\bottomrule
\end{tabular}

\end{table}

\paragraph{Scope and reach of the certificate condition.}
How much repair work certificate provenance saves depends on certificate size relative to fan-in. An absorbing input yields a certificate of size one (e.g., a false input to $\land$ or a true input to $\lor$), and a $k$-of-$n$ threshold rule admits a certificate of size $k$ (or $n-k+1$). Conversely, a numeric sum or Boolean parity requires all parents in every certificate, leaving no room for selectivity, while a numeric maximum requires all parents unless one parent attains the maximum possible domain value. Notice also that certificate size governs whether structural edges can be omitted at formation ($C_v\subsetneq\mathrm{Pa}(v)$), whereas whether a descendant actually changes value under a specific update ($\Asens(u,t)$) also depends on the update and path structure. Certificate reachability is a sound over-approximation of update sensitivity, not an equivalence: even when $C_v=\mathrm{Pa}(v)$, reconvergent paths can cancel out so that a structural descendant's value does not change (for example, if $a=u$, $b=u$, and $v=a\oplus b$, the unique certificate of $v$ at $(0,0)$ is its full parent set $\{a,b\}$, yet flipping $u:0\to 1$ flips both $a$ and $b$ and leaves $v=0$ unchanged, so $v\in\Astruct(u,t)\setminus\Asens(u,t)$). Full-parent certificates imply $\Astruct=\Asens$ when, in addition, dependencies follow a single non-canceling path whose hops are injective on the updated value, as in single-path relational lookup chains. For set-valued or multi-claim text summaries, each atomic claim requires its own certificate; partial updates then invalidate only the claims whose certificates touch the updated source when claims are stored modularly, whereas a monolithic text entry requires the union of certificates across all its constituent claims.

\paragraph{Formation-time versus audit-time sensitivity.}
Under oracle structural closure, every audit of an ancestor re-derives the descendant, so formation-time and audit-time value-agreement edges coincide exactly; across LLM-provenance trajectories they differ on $2.7$--$4.5\%$ of stored pairs. Similarly, evaluating single-root sensitivity only at formation time assigns no sensitive root to $104/737$ ($14.1\%$) of lifecycle-stale descendants in the depth sweep ($36/288$, or $12.5\%$, in the base setting) because those nodes require two or more sequential corrections before their task value flips. Evaluating $\Asens(u,t)$ at audit time $t$ via Eq.~(2) resolves this discrepancy.

\paragraph{Sensitivity, actual staleness, and out-of-lineage reader errors.}
Table~\ref{tab:app_three_set} reports exact pair counts for $\Astruct$, $\Asens$, and $\Astale$, along with pre-existing reader errors outside the corrected root's lineage ($67$ Luna root--state pairs in the depth sweep and $34$ in the base setting; $0$ for DeepSeek). Structural coverage gaps remain unchanged across oracle, LLM-active, and pre-correction denominators ($0.559$--$0.590$ pooled over depth), and condition ordering is invariant between LOW and HIGH error placements (rank correlation $\approx 0.99$).

\begin{table}[!htbp]
\caption{Counts of structural ($\Astruct$), audit-time sensitive ($\Asens$), and post-audit stale ($\Astale$) root--descendant pairs over active states in oracle and LLM reachability replays. Stale-but-insensitive pairs are partitioned by cause: formation-time reader error versus missed earlier repair ($0$ in all conditions).}
\label{tab:app_three_set}
\centering
\scriptsize
\resizebox{\linewidth}{!}{\begin{tabular}{lllrrrrrrr}
\toprule
Setting & Reader & Replay graph & $|A_{\rm struct}|$ & $|A_{\rm sens}|$ & $|A_{\rm stale}|$ & both & sens$\setminus$stale & stale$\setminus$sens (reader err.\ / earlier audit) & out-of-lineage stale \\
\midrule
depth sweep D1–5 & Luna & oracle & 2047 & 614 & 608 & 597 & 17 & 11 / 0 & 55 \\
depth sweep D1–5 & Luna & llm & 2651 & 682 & 681 & 664 & 18 & 17 / 0 & 67 \\
depth sweep D1–5 & DeepSeek & oracle & 2050 & 621 & 621 & 621 & 0 & 0 / 0 & 0 \\
depth sweep D1–5 & DeepSeek & llm & 2630 & 682 & 682 & 682 & 0 & 0 / 0 & 0 \\
main setting & Luna & oracle & 864 & 251 & 252 & 248 & 3 & 4 / 0 & 34 \\
main setting & Luna & llm & 997 & 274 & 275 & 271 & 3 & 4 / 0 & 34 \\
main setting & DeepSeek & oracle & 864 & 251 & 251 & 251 & 0 & 0 / 0 & 0 \\
main setting & DeepSeek & llm & 997 & 274 & 274 & 274 & 0 & 0 / 0 & 0 \\
\bottomrule
\end{tabular}}
\end{table}

\FloatBarrier
\section{Public Benchmark Details, Unit Accounting, and LLM Attribution}
\label{app:public}

MemoryAgentBench Conflict Resolution / FactConsolidation \citep{hu2026memoryagentbench} provides long-context streams of timestamped facts and multi-hop questions. A later fact with the same subject--relation key updates an earlier value. For each multi-hop question, we issue an in-stream query once all facts along the pre-correction path have arrived, persist the intermediate hop entity in memory, and probe the final question after all subsequent corrections conclude. Unlike the controlled Boolean testbed where the dependency DAG is fixed throughout the episode, updating an intermediate entity in a relational lookup chain dynamically changes which downstream fact is queried on the next hop; thus, MemoryAgentBench serves as an empirical generalization test under dynamic dependencies rather than a direct instantiation of Proposition~\ref{prop:certificate}'s fixed-DAG model.

\paragraph{Unit selection and component-clustered confidence intervals.}
An evaluation unit is a multi-hop question possessing a unique relation path whose pre-correction path is complete at the intermediate query time and differs from the final post-correction path at some hop. Reversing the temporal order of initial and corrected values changes which questions have a complete pre-correction path window before the update arrives: for instance, in \texttt{mh\_6k}, $4$ questions lack a resolved path in both orderings, while $29$ questions in the original order versus $34$ in the reversed order lack a pre-correction stale window, yielding $67$ versus $62$ evaluated units (Table~\ref{tab:units_clustering}). Because all units within a split are drawn from a shared fact stream, stream-level resampling is not defined; to account for shared facts across questions, we group units that share any subject--relation key into connected components ($30$--$46$ components per split) and compute component-clustered bootstrap 95\% CIs alongside unit-clustered CIs. Seven of eight write-back $\times$ validity interactions have component-clustered 95\% CIs strictly below zero. The exception is \texttt{mh\_32k}, reversed order, DeepSeek: its interval is $[-0.19,0.00]$, with an upper endpoint exactly equal to zero (Table~\ref{tab:units_clustering}).

\begin{table}[!htbp]
\caption{Public benchmark unit accounting and comparison of unit-clustered versus connected-component-clustered bootstrap 95\% confidence intervals for the write-back $\times$ validity interaction.}
\label{tab:units_clustering}
\centering
\scriptsize
\resizebox{\linewidth}{!}{\begin{tabular}{llccccccc}
\toprule
Split & Order & Kept units & No path & No stale window & Fact components & Reader & Interaction, unit CI & Interaction, component CI \\
\midrule
mh\_6k & reversed & 62 & 4 & 34 & 35 & Luna & -0.37 [-0.52, -0.22] & [-0.62, -0.21] \\
 &  &  &  &  &  & DeepSeek & -0.25 [-0.41, -0.11] & [-0.38, -0.09] \\
mh\_6k & original & 67 & 4 & 29 & 46 & Luna & -0.86 [-1.01, -0.71] & [-1.05, -0.74] \\
 &  &  &  &  &  & DeepSeek & -0.95 [-1.00, -0.86] & [-1.00, -0.88] \\
mh\_32k & reversed & 52 & 16 & 31 & 30 & Luna & -0.34 [-0.53, -0.17] & [-0.53, -0.10] \\
 &  &  &  &  &  & DeepSeek & -0.07 [-0.17, +0.00] & [-0.19, +0.00] \\
mh\_32k & original & 65 & 16 & 18 & 42 & Luna & -0.84 [-0.98, -0.68] & [-0.99, -0.68] \\
 &  &  &  &  &  & DeepSeek & -0.90 [-1.00, -0.79] & [-1.00, -0.79] \\
\bottomrule
\end{tabular}
}
\end{table}

\begin{table}[!htbp]
\caption{LLM attribution versus oracle path support across eight MemoryAgentBench cells. Accuracy includes all evaluated units, both valid and invalid intermediates: \texttt{mh\_6k} has $67/62$ units in original/reversed order and \texttt{mh\_32k} has $65/52$. Accuracy differences (\texttt{LLM} $-$ \texttt{Oracle}) show component-clustered 95\% CIs.}
\label{tab:public_llm}
\centering
\scriptsize
\resizebox{\linewidth}{!}{\begin{tabular}{lllcccccccc}
\toprule
Split & Order & Reader & No-write (\texttt{OFF}) & Keep-stale (\texttt{S0}) & Oracle & LLM & LLM$-$oracle & Struct. cov. & Stale cov. & Edge P / R \\
\midrule
mh\_6k & reversed & Luna & 0.935 & 0.710 & 0.935 & 0.935 & +0.000 [+0.000, +0.000] & 0.75 & 0.75 [0.61, 0.88] & 0.99 / 0.88 \\
mh\_6k & reversed & DeepSeek & 0.968 & 0.855 & 0.968 & 0.968 & +0.000 [+0.000, +0.000] & 0.89 & 0.89 [0.79, 0.97] & 0.99 / 0.93 \\
mh\_6k & original & Luna & 0.955 & 0.478 & 0.955 & 0.955 & +0.000 [+0.000, +0.000] & 0.96 & 0.96 [0.89, 1.00] & 0.99 / 0.93 \\
mh\_6k & original & DeepSeek & 1.000 & 0.478 & 1.000 & 1.000 & +0.000 [+0.000, +0.000] & 0.98 & 0.98 [0.92, 1.00] & 1.00 / 0.96 \\
mh\_32k & reversed & Luna & 0.962 & 0.769 & 0.962 & 0.962 & +0.000 [+0.000, +0.000] & 0.85 & 0.88 [0.75, 0.97] & 0.98 / 0.92 \\
mh\_32k & reversed & DeepSeek & 0.981 & 0.942 & 0.981 & 0.981 & +0.000 [+0.000, +0.000] & 0.88 & 0.87 [0.74, 0.97] & 0.99 / 0.94 \\
mh\_32k & original & Luna & 0.862 & 0.492 & 0.862 & 0.862 & +0.000 [+0.000, +0.000] & 0.92 & 0.92 [0.82, 1.00] & 0.97 / 0.89 \\
mh\_32k & original & DeepSeek & 0.985 & 0.554 & 0.969 & 0.985 & +0.015 [+0.000, +0.040] & 0.97 & 0.97 [0.91, 1.00] & 0.99 / 0.93 \\
\bottomrule
\end{tabular}
}
\end{table}

\begin{figure}[!htbp]
  \centering
  \includegraphics[width=\linewidth]{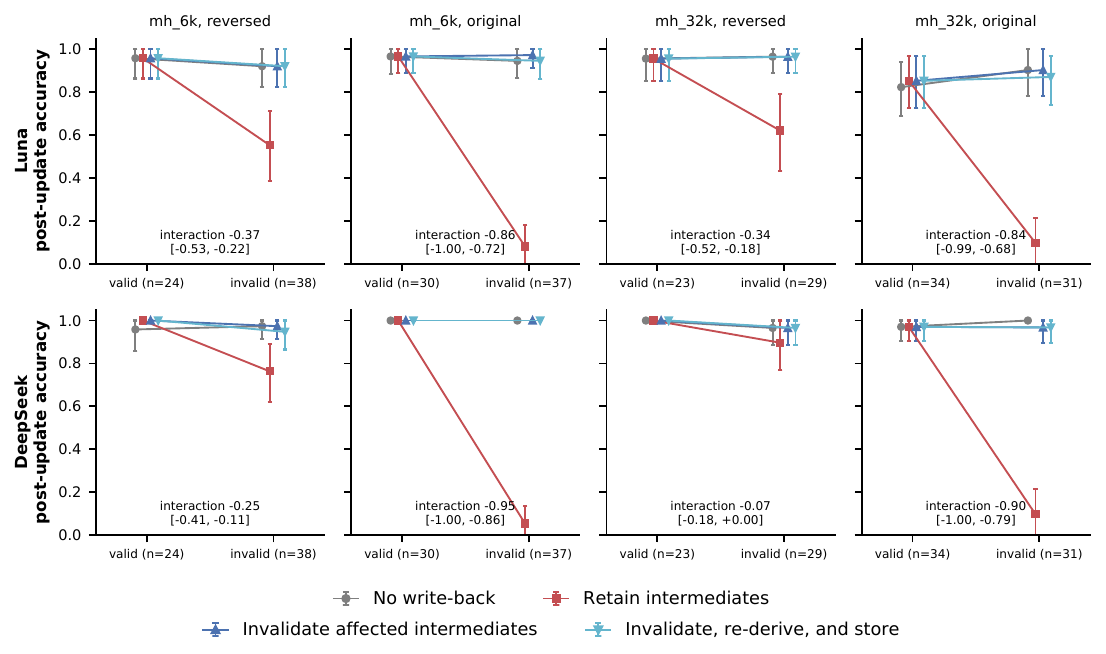}
  \caption{\textbf{Full public multi-hop validation grid under oracle path support.} The four policies are no write-back, retention of intermediates, closure invalidation, and invalidation followed by re-derivation and write-back.}
  \label{fig:app_public_full}
\end{figure}

\begin{table}[!htbp]
\caption{Public write-back $\times$ validity interactions across all eight cells. Negative values indicate that persistence is more harmful when the stored intermediate becomes invalid.}
\label{tab:app_public}
\centering
\scriptsize
\resizebox{\linewidth}{!}{\begin{tabular}{lllcccccc}
\toprule
Split & Order & Reader & $n$ valid / invalid & No-write (\texttt{OFF}, inv.) & Keep-stale (\texttt{S0}, inv.) & Invalidate (\texttt{S4}, inv.) & Interaction [95\% CI] & $b{:}c$ \\
\midrule
mh\_32k & reversed & Luna & 23 / 29 & 0.966 & 0.621 & 0.966 & -0.34 [-0.52, -0.18] & 10:0 \\
mh\_32k & reversed & DeepSeek & 23 / 29 & 0.966 & 0.897 & 0.966 & -0.07 [-0.18, 0.00] & 2:0 \\
mh\_32k & original & Luna & 34 / 31 & 0.903 & 0.097 & 0.903 & -0.84 [-0.99, -0.68] & 25:0 \\
mh\_32k & original & DeepSeek & 34 / 31 & 1.000 & 0.097 & 0.968 & -0.90 [-1.00, -0.79] & 28:0 \\
mh\_6k & reversed & Luna & 24 / 38 & 0.921 & 0.553 & 0.921 & -0.37 [-0.53, -0.22] & 14:0 \\
mh\_6k & reversed & DeepSeek & 24 / 38 & 0.974 & 0.763 & 0.974 & -0.25 [-0.41, -0.11] & 8:0 \\
mh\_6k & original & Luna & 30 / 37 & 0.946 & 0.081 & 0.973 & -0.86 [-1.00, -0.72] & 32:0 \\
mh\_6k & original & DeepSeek & 30 / 37 & 1.000 & 0.054 & 1.000 & -0.95 [-1.00, -0.86] & 35:0 \\
\bottomrule
\end{tabular}
}
\end{table}

\begin{figure}[!htbp]
  \centering
  \includegraphics[width=0.88\linewidth]{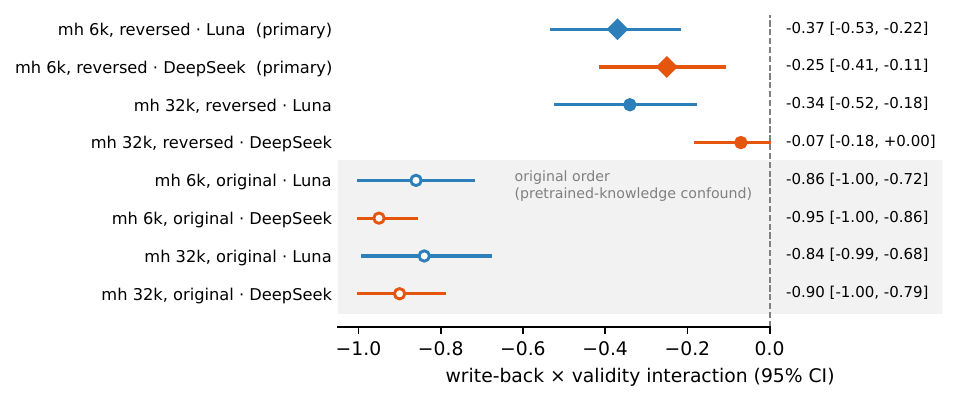}
  \caption{\textbf{Forest plot of write-back $\times$ validity interactions across all public cells} (Table~\ref{tab:app_public}). Filled diamonds denote the primary reversed short-context control; open circles denote original-order splits.}
  \label{fig:app_forest}
\end{figure}

\FloatBarrier
\section{Evaluation Counts and Graph Topology Statistics}
\label{app:counts}

Tables~\ref{tab:eval_counts} and~\ref{tab:graph_stats} report the evaluation unit counts and the observed path depth, fan-in, and fan-out statistics across all controlled and public benchmark settings.

\begin{table}[!htbp]
\caption{Summary of evaluation unit counts across controlled and public analyses.}
\label{tab:eval_counts}
\centering
\small
\begin{tabular}{lc}
\toprule
Evaluation setting & Count \\
\midrule
Controlled base episodes & 50 \\
Controlled depth variants ($D=1$--$5$) & $50\times5$ \\
Closed-loop depth variants ($D\in\{1,3,5\}$) & $50\times3$ \\
Source claims (roots) / episode & 24 (6 incorrect) \\
Base in-stream derived queries / episode & 9 \\
Public single-hop validation units (6k+32k) & 137 \\
Public multi-hop 6k units (original / reversed) & 67 / 62 \\
Public multi-hop 32k units (original / reversed) & 65 / 52 \\
Primary reversed 6k valid / invalid units & 24 / 38 \\
\bottomrule
\end{tabular}
\end{table}

\begin{table}[!htbp]
\caption{Observed graph depth, fan-in (parents per derived node), and fan-out (children per node) across controlled and public settings.}
\label{tab:graph_stats}
\centering
\small
\begin{tabular}{lccccc}
\toprule
Setting & Max depth & Max fan-in & Mean fan-in & Max fan-out & Mean fan-out \\
\midrule
depth & 5 & 4 & 3.24 & 6 & 1.83 \\
nonunate & 2 & 4 & 3.20 & 4 & 1.78 \\
wide & 4 & 4 & 3.30 & 10 & 2.35 \\
public mh\_6k original & 3 & 3 & 1.36 & -- & -- \\
public mh\_6k reversed & 3 & 3 & 1.34 & -- & -- \\
public mh\_32k original & 3 & 3 & 1.45 & -- & -- \\
public mh\_32k reversed & 3 & 3 & 1.35 & -- & -- \\
\bottomrule
\end{tabular}

\end{table}

\FloatBarrier
\section{Prior Repair Policies in the Controlled Testbed}
\label{app:prior}

Because prior agent-memory repair papers either do not release code (\textsc{MemoRepair}, \textsc{Revise}) or target tool-execution traces rather than rule-governed memory (\citealt{yu2026rollback}), we instantiate each system's core repair decision rule within our controlled lifecycle while keeping episodes, readers, prompts, and the re-derivation operator identical. Table~\ref{tab:prior_deviations} documents the exact adaptations, and Table~\ref{tab:closed_loop} reports the closed-loop results alongside our provenance conditions.

\begin{table}[!htbp]
\centering
\scriptsize
\caption{Instantiation of prior repair policies in the controlled testbed.}
\label{tab:prior_deviations}
\begin{tabular}{p{2.5cm}p{4.8cm}p{5.4cm}}
\toprule
Policy & Preserved from source work & Adapted or omitted for this testbed \\
\midrule
\textsc{MemoRepair} \citep{zhao2026memorepair} & Barrier withdrawal of the reachable closure; predecessor-closed publication selection $\max \sum_i w_i x_i-\lambda\sum_i c_i x_i$ solved via min-cut with default $\lambda=0.3$; topological republication. & Uniform utility $w_i=1$; cost $c_i$ set to measured re-derivation tokens normalized by mean; schema validity check; no neural-skill or tool artifacts. Under these parameters the selector retains $2{,}659/2{,}659$ candidates and reduces to oracle structural closure. \\
Withdraw only & \textsc{MemoRepair}'s ``remove all'' baseline: withdraw reachable closure without republication. & Same reachability closure as above. \\
\textsc{Revise} \citep{qi2026revise} & Recompute affected subgraph from recorded reads; expand recovery when provenance is incomplete. & With complete recorded reads at $K=2$, the read graph equals $\Gtrue$, coinciding with oracle closure. \\
Support check \citep{yu2026rollback} & Preserve a reached node if it retains at least one direct parent outside the faulty and affected sets (non-recursive lenient check). & Omitted fault-specific quarantine and final-answer replay; applied to conjunctive/threshold task rules rather than independent evidential paths. \\
Early cutoff \citep{acar2002adaptive,hammer2014adapton} & Re-derive direct children of corrected roots in topological order; propagate further only when a re-derived node's output changes. & Requires full oracle structural edges $\Gtrue$ to identify children of changed nodes. \\
\bottomrule
\end{tabular}
\end{table}

\begin{table}[!htbp]
\caption{Closed-loop (end-to-end) comparison of provenance representations and repair policies across $D\in\{1,3,5\}$ (50 episodes per reader). Every repair regenerates answers via the reader LLM and refreshes attribution.}
\label{tab:closed_loop}
\centering
\scriptsize
\resizebox{\linewidth}{!}{\begin{tabular}{llcccccccc}
\toprule
Reader & Provenance / policy & Edges/node & Edge R & Local corr. & Cond.\ cert. & Stale cov. & Residual stale & Probe err./ep. & Re-deriv./ep. \\
\midrule
Luna & No repair & -- & -- & -- & -- & -- & -- & 0.53 [0.36, 0.72] & 0.0 \\
 & Withdraw only, no re-derivation & -- & -- & -- & -- & -- & -- & 0.71 [0.53, 0.89] & 0.0 \\
 & Support check (adapted from rollback repair) & 3.50 & 1.00 & 0.987 & 1.000 & 0.00 [0.00, 0.00] & 1.000 [1.000, 1.000] & 0.53 [0.36, 0.72] & 0.0 \\
 & Rule-based but-for & 1.51 & 0.43 & 0.990 & 0.715 & 0.83 [0.74, 0.91] & 0.162 [0.081, 0.248] & 0.14 [0.04, 0.26] & 2.8 \\
 & Matched random & 3.04 & 0.71 & 0.992 & 0.625 & 0.64 [0.55, 0.73] & 0.260 [0.188, 0.340] & 0.21 [0.11, 0.31] & 6.6 \\
 & Minimal certificate & 1.90 & 0.55 & 0.989 & 1.000 & 0.98 [0.94, 1.00] & 0.015 [0.000, 0.039] & 0.04 [0.00, 0.10] & 4.6 \\
 & Value-agreement & 2.30 & 0.68 & 0.990 & 1.000 & 0.97 [0.94, 1.00] & 0.016 [0.000, 0.038] & 0.05 [0.00, 0.11] & 5.7 \\
 & LLM material-support (deepseek-flash) & 2.96 & 0.69 & 0.990 & 0.9995 & 0.97 [0.95, 1.00] & 0.014 [0.000, 0.032] & 0.05 [0.00, 0.11] & 6.0 \\
 & LLM material-support (gpt-5.6-luna) & 3.26 & 0.79 & 0.990 & 0.998 & 0.94 [0.90, 0.98] & 0.036 [0.012, 0.065] & 0.07 [0.01, 0.15] & 7.4 \\
 & Oracle structural + early cutoff & 3.56 & 1.00 & 0.991 & 1.000 & 0.98 [0.96, 1.00] & 0.012 [0.000, 0.030] & 0.04 [0.00, 0.10] & 6.8 \\
 & Oracle structural closure & 3.23 & 1.00 & 0.993 & 1.000 & 1.00 [1.00, 1.00] & 0.000 [0.000, 0.000] & 0.02 [0.00, 0.06] & 10.8 \\
\midrule
DeepSeek & No repair & -- & -- & -- & -- & -- & -- & 0.47 [0.31, 0.65] & 0.0 \\
 & Withdraw only, no re-derivation & -- & -- & -- & -- & -- & -- & 0.65 [0.48, 0.83] & 0.0 \\
 & Support check (adapted from rollback repair) & 3.50 & 1.00 & 1.000 & 1.000 & 0.00 [0.00, 0.00] & 1.000 [1.000, 1.000] & 0.47 [0.31, 0.65] & 0.0 \\
 & Rule-based but-for & 1.51 & 0.43 & 1.000 & 0.717 & 0.83 [0.75, 0.91] & 0.151 [0.071, 0.242] & 0.12 [0.02, 0.25] & 2.7 \\
 & Matched random & 3.08 & 0.72 & 1.000 & 0.622 & 0.66 [0.58, 0.75] & 0.248 [0.176, 0.328] & 0.19 [0.09, 0.30] & 6.7 \\
 & Minimal certificate & 1.91 & 0.56 & 1.000 & 1.000 & 1.00 [1.00, 1.00] & 0.000 [0.000, 0.000] & 0.00 [0.00, 0.00] & 4.7 \\
 & Value-agreement & 2.31 & 0.68 & 1.000 & 1.000 & 1.00 [1.00, 1.00] & 0.000 [0.000, 0.000] & 0.00 [0.00, 0.00] & 5.8 \\
 & LLM material-support (deepseek-flash) & 2.98 & 0.69 & 1.000 & 0.9981 & 1.00 [1.00, 1.00] & 0.000 [0.000, 0.000] & 0.00 [0.00, 0.00] & 6.0 \\
 & LLM material-support (gpt-5.6-luna) & 3.26 & 0.79 & 1.000 & 0.9983 & 1.00 [1.00, 1.00] & 0.000 [0.000, 0.000] & 0.00 [0.00, 0.00] & 7.6 \\
 & Oracle structural + early cutoff & 3.56 & 1.00 & 1.000 & 1.000 & 1.00 [1.00, 1.00] & 0.000 [0.000, 0.000] & 0.00 [0.00, 0.00] & 6.8 \\
 & Oracle structural closure & 3.23 & 1.00 & 1.000 & 1.000 & 1.00 [1.00, 1.00] & 0.000 [0.000, 0.000] & 0.00 [0.00, 0.00] & 10.8 \\
\bottomrule
\end{tabular}
}
\end{table}

Two findings stand out from Table~\ref{tab:closed_loop}:
\begin{enumerate}[leftmargin=1.8em,itemsep=1pt]
    \item When supplied with complete structural provenance, \textsc{MemoRepair} and \textsc{Revise} invalidate the same closure as oracle structural repair ($10.8$ re-derivations per episode). Evaluating our LLM material-support graph by \textsc{MemoRepair}'s $\text{Leak\%}$ metric ($1-C_{\mathrm{struct}}$) would penalize it for ``leaking'' $44$--$45\%$ of the structural cascade, even though it repairs $97$--$100\%$ of stale state and achieves $0.05/0.00$ probe errors with $6.0$ re-derivations.
    \item The lenient support check of \citet{yu2026rollback} keeps any derived node that has at least one untouched parent. Because L1 nodes have four parents and L2 nodes have two, a single-root audit leaves untouched parents at every node, so the lenient check re-derives nothing ($0.0$ re-derivations) and leaves probe errors identical to no repair ($0.53/0.47$). This reflects the difference between independent evidential witnesses and multi-input logical rules. Meanwhile, incremental early cutoff on $\Gtrue$ achieves near-oracle accuracy ($0.04/0.00$ errors) with $6.8$ re-derivations, which is still higher than certificate ($4.6$--$4.7$) or value-agreement ($5.7$--$5.8$) repair because early cutoff must always re-evaluate all direct children of a changed node before halting.
\end{enumerate}

\FloatBarrier
\section{Retrieval Sensitivity, Edge Corruption, and Stress Families}
\label{app:sensitivity}

\begin{figure}[H]
  \centering
  \includegraphics[width=\linewidth]{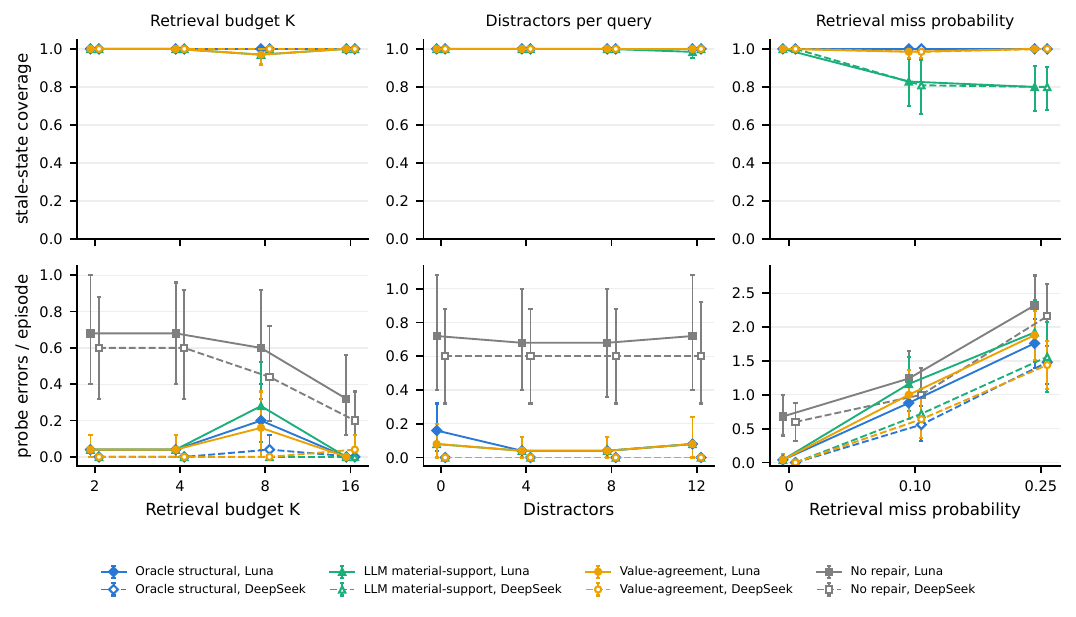}
  \caption{\textbf{Closed-loop sensitivity to retrieval budget $K$, distractor count, and retrieval miss probability} (Table~\ref{tab:sensitivity}). Budget and distractor variations preserve near-complete stale coverage, whereas missing rule-required parents during retrieval degrades all methods and lowers LLM stale coverage.}
  \label{fig:sensitivity}
\end{figure}

\paragraph{Sensitivity to retrieval budget, distractors, and retrieval misses.}
Table~\ref{tab:sensitivity} and Figure~\ref{fig:sensitivity} report closed-loop sensitivity sweeps on the base setting (LOW placement, 25 episodes per reader), varying the L2 retrieval budget $K\in\{2,4,8,16\}$, the number of distractors per query ($0,4,8,12$), and the probability $p_{\mathrm{miss}}\in\{0,0.10,0.25\}$ that each rule-required parent memory is omitted during retrieval. Varying $K$ and distractor counts leaves stale-state coverage of LLM and value-agreement provenance at $0.97$--$1.00$. By contrast, when required parent memories are randomly missed during retrieval ($p_{\mathrm{miss}}=0.25$), every method suffers high probe error---including oracle structural closure ($1.76/1.48$ errors per episode)---because the reader must guess without required inputs, and LLM stale-state coverage drops to $0.80/0.80$ as the attributor occasionally cites candidate memories that did not actually determine the guessed answer.

\paragraph{False-positive and false-negative provenance edges.}
Table~\ref{tab:fpfn} evaluates closed-loop runs where oracle edges are randomly dropped (false negatives) or supplemented with spurious edges (false positives). Adding spurious edges (50\% random spurious edges or 4 context co-occurrence edges per node) keeps stale-state coverage at $1.00$ while increasing re-derivations from $10.0$ to $13.5$--$18.1$ per episode. Dropping true edges reduces stale-state coverage in direct proportion to certificate containment (see also Fig.~\ref{fig:closed_loop}C).

\begin{figure}[H]
  \centering
  \includegraphics[width=\linewidth]{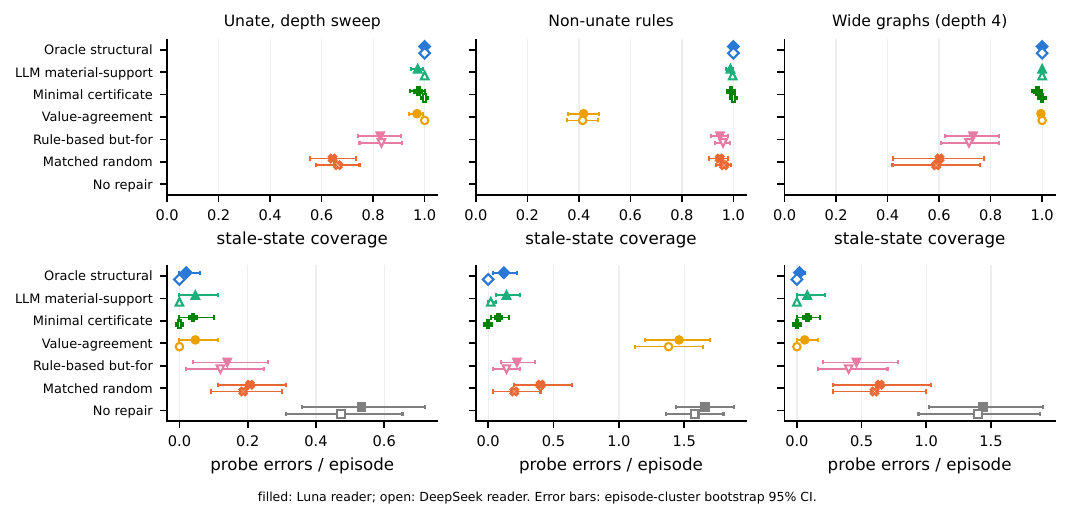}
  \caption{\textbf{Closed-loop performance on non-unate rules and wide branching graphs} (Table~\ref{tab:stress}). On non-unate rules, value agreement fails as predicted while minimal certificates and LLM attribution remain near oracle; on wide graphs, the separation between selective provenance and matched random widens.}
  \label{fig:stress}
\end{figure}

\paragraph{Non-unate rules and wide branching graphs.}
Table~\ref{tab:stress} and Figure~\ref{fig:stress} evaluate two closed-loop stress families reusing the same frozen roots and audit schedules:
\begin{itemize}[leftmargin=1.5em,itemsep=1pt]
    \item \textbf{Non-unate family:} L1 rules are four-input parity, exactly-two-of-four, and $(A\land B)\lor(\lnot A\land C)\lor D$; L2 rules are XOR and logical equivalence. As predicted by theory, value agreement is no longer a certificate here and reaches only $0.42/0.41$ of stale state ($1.46/1.38$ probe errors per episode). By contrast, minimal certificates reach $0.99/1.00$ of stale state ($0.08/0.00$ errors), and LLM material-support attribution also reaches $0.99/1.00$ ($0.14/0.02$ errors vs.\ oracle $0.12/0.00$) by retaining more parents on non-unate rules (structural coverage $0.84/0.85$ vs.\ $0.40/0.39$ for value agreement). This confirms that LLM attribution tracks certificate sufficiency rather than simple sign agreement when rules are non-monotone.
    \item \textbf{Wide branching family (depth 4, fan-in up to 4, shared parents, max fan-out 10):} On wider graphs with heavier parent sharing, LLM provenance and value agreement both reach $1.00/1.00$ of stale state while covering only $0.49$ and $0.45$ of structural descendants ($12.5$--$12.6$ vs.\ $25.8$ oracle re-derivations per episode). Matched random with identical edge budgets reaches only $0.60/0.59$ of stale state and leaves $0.64/0.60$ probe errors per episode versus $0.08/0.00$ for LLM provenance.
\end{itemize}

\begin{table}[H]
\caption{Closed-loop sensitivity to retrieval budget $K$, distractors, and retrieval misses of rule-required memories (25 episodes per reader).}
\label{tab:sensitivity}
\centering
\scriptsize
\resizebox{\linewidth}{!}{\begin{tabular}{llcccccc}
\toprule
Reader & Setting & Stale cov.\ (LLM) & Stale cov.\ (value) & Probe err.\ (no repair) & Oracle & LLM & Value \\
\midrule
Luna & K=2, 4 distractors, no misses & 1.00 [1.00, 1.00] & 1.00 [1.00, 1.00] & 0.68 & 0.04 & 0.04 & 0.04 \\
 & K=4 & 1.00 [1.00, 1.00] & 1.00 [1.00, 1.00] & 0.68 & 0.04 & 0.04 & 0.04 \\
 & K=8 & 0.97 [0.92, 1.00] & 0.97 [0.92, 1.00] & 0.60 & 0.20 & 0.28 & 0.16 \\
 & K=16 & 1.00 [1.00, 1.00] & 1.00 [1.00, 1.00] & 0.32 & 0.00 & 0.00 & 0.00 \\
 & 0 distractors & 1.00 [1.00, 1.00] & 1.00 [1.00, 1.00] & 0.72 & 0.16 & 0.08 & 0.08 \\
 & 8 distractors & 1.00 [1.00, 1.00] & 1.00 [1.00, 1.00] & 0.68 & 0.04 & 0.04 & 0.04 \\
 & 12 distractors & 0.98 [0.95, 1.00] & 1.00 [1.00, 1.00] & 0.72 & 0.08 & 0.08 & 0.08 \\
 & retrieval miss 0.10 & 0.83 [0.70, 0.95] & 0.98 [0.95, 1.00] & 1.24 & 0.88 & 1.16 & 1.00 \\
 & retrieval miss 0.25 & 0.80 [0.67, 0.91] & 1.00 [1.00, 1.00] & 2.32 & 1.76 & 1.92 & 1.88 \\
\midrule
DeepSeek & K=2, 4 distractors, no misses & 1.00 [1.00, 1.00] & 1.00 [1.00, 1.00] & 0.60 & 0.00 & 0.00 & 0.00 \\
 & K=4 & 1.00 [1.00, 1.00] & 1.00 [1.00, 1.00] & 0.60 & 0.00 & 0.00 & 0.00 \\
 & K=8 & 1.00 [1.00, 1.00] & 1.00 [1.00, 1.00] & 0.44 & 0.04 & 0.00 & 0.00 \\
 & K=16 & 1.00 [1.00, 1.00] & 1.00 [1.00, 1.00] & 0.20 & 0.00 & 0.00 & 0.04 \\
 & 0 distractors & 1.00 [1.00, 1.00] & 1.00 [1.00, 1.00] & 0.60 & 0.00 & 0.00 & 0.00 \\
 & 8 distractors & 1.00 [1.00, 1.00] & 1.00 [1.00, 1.00] & 0.60 & 0.00 & 0.00 & 0.00 \\
 & 12 distractors & 1.00 [1.00, 1.00] & 1.00 [1.00, 1.00] & 0.60 & 0.00 & 0.00 & 0.00 \\
 & retrieval miss 0.10 & 0.81 [0.66, 0.94] & 0.98 [0.95, 1.00] & 1.00 & 0.56 & 0.72 & 0.64 \\
 & retrieval miss 0.25 & 0.80 [0.68, 0.90] & 1.00 [1.00, 1.00] & 2.16 & 1.48 & 1.56 & 1.44 \\
\bottomrule
\end{tabular}
}
\end{table}

\begin{table}[H]
\caption{Closed-loop evaluation under controlled false-negative (edge drop) and false-positive (spurious edge addition) provenance perturbations.}
\label{tab:fpfn}
\centering
\scriptsize
\resizebox{\linewidth}{!}{\begin{tabular}{llccccccc}
\toprule
Reader & Perturbation condition & Edge P & Edge R & Local corr. & Cond.\ cert. & Stale cov. & Probe err./ep. & Re-deriv./ep. \\
\midrule
Luna & Oracle ($G^{\star}$) & 1.00 & 1.00 & 0.995 & 1.00 & 1.00 [1.00, 1.00] & 0.04 [0.00, 0.09] & 10.0 \\
 & Drop 10\% true edges & 1.00 & 0.90 & 0.995 & 0.86 & 0.85 [0.80, 0.90] & 0.17 [0.09, 0.26] & 8.4 \\
 & Drop 25\% true edges & 1.00 & 0.74 & 0.995 & 0.64 & 0.60 [0.53, 0.68] & 0.43 [0.29, 0.58] & 6.5 \\
 & Drop 50\% true edges & 1.00 & 0.51 & 0.995 & 0.37 & 0.37 [0.31, 0.43] & 0.56 [0.40, 0.72] & 4.1 \\
 & Add 25\% spurious edges & 0.84 & 1.00 & 0.995 & 1.00 & 1.00 [1.00, 1.00] & 0.02 [0.00, 0.05] & 13.7 \\
 & Add 50\% spurious edges & 0.67 & 1.00 & 0.995 & 1.00 & 1.00 [1.00, 1.00] & 0.01 [0.00, 0.03] & 18.1 \\
 & Drop 25\% true + add 25\% spurious & 0.79 & 0.74 & 0.995 & 0.64 & 0.62 [0.54, 0.69] & 0.38 [0.25, 0.53] & 9.1 \\
 & Drop 25\% true + add 50\% spurious & 0.60 & 0.74 & 0.995 & 0.65 & 0.64 [0.57, 0.71] & 0.28 [0.19, 0.38] & 12.5 \\
 & Drop 50\% non-cited true edges & 1.00 & 0.80 & 0.995 & 0.76 & 0.75 [0.69, 0.81] & 0.20 [0.12, 0.29] & 7.6 \\
 & Drop 100\% non-cited true edges & 1.00 & 0.58 & 0.995 & 0.63 & 0.40 [0.34, 0.47] & 0.56 [0.41, 0.73] & 5.3 \\
 & Depth-decayed recall (0.9 / 0.6) & 1.00 & 0.84 & 0.995 & 0.77 & 0.79 [0.74, 0.83] & 0.24 [0.15, 0.34] & 7.0 \\
 & Context co-occurrence (+4 spurious) & 0.73 & 1.00 & 0.995 & 1.00 & 1.00 [1.00, 1.00] & 0.02 [0.00, 0.05] & 13.5 \\
\midrule
DeepSeek & Oracle ($G^{\star}$) & 1.00 & 1.00 & 1.000 & 1.00 & 1.00 [1.00, 1.00] & 0.00 [0.00, 0.00] & 10.0 \\
 & Drop 10\% true edges & 1.00 & 0.90 & 1.000 & 0.86 & 0.85 [0.80, 0.90] & 0.13 [0.07, 0.20] & 8.4 \\
 & Drop 25\% true edges & 1.00 & 0.74 & 1.000 & 0.64 & 0.61 [0.53, 0.69] & 0.37 [0.25, 0.51] & 6.5 \\
 & Drop 50\% true edges & 1.00 & 0.51 & 1.000 & 0.37 & 0.37 [0.31, 0.44] & 0.52 [0.37, 0.68] & 4.1 \\
 & Add 25\% spurious edges & 0.84 & 1.00 & 1.000 & 1.00 & 1.00 [1.00, 1.00] & 0.00 [0.00, 0.00] & 13.7 \\
 & Add 50\% spurious edges & 0.67 & 1.00 & 1.000 & 1.00 & 1.00 [1.00, 1.00] & 0.00 [0.00, 0.00] & 18.1 \\
 & Drop 25\% true + add 25\% spurious & 0.79 & 0.74 & 1.000 & 0.64 & 0.62 [0.54, 0.70] & 0.33 [0.21, 0.46] & 9.1 \\
 & Drop 25\% true + add 50\% spurious & 0.60 & 0.74 & 1.000 & 0.65 & 0.65 [0.59, 0.72] & 0.25 [0.16, 0.34] & 12.5 \\
 & Drop 50\% non-cited true edges & 1.00 & 0.80 & 1.000 & 0.77 & 0.76 [0.70, 0.81] & 0.16 [0.08, 0.24] & 7.6 \\
 & Drop 100\% non-cited true edges & 1.00 & 0.58 & 1.000 & 0.63 & 0.41 [0.35, 0.48] & 0.49 [0.35, 0.63] & 5.3 \\
 & Depth-decayed recall (0.9 / 0.6) & 1.00 & 0.84 & 1.000 & 0.77 & 0.79 [0.74, 0.84] & 0.20 [0.12, 0.29] & 7.0 \\
 & Context co-occurrence (+4 spurious) & 0.73 & 1.00 & 1.000 & 1.00 & 1.00 [1.00, 1.00] & 0.00 [0.00, 0.00] & 13.5 \\
\bottomrule
\end{tabular}
}
\end{table}

\begin{table}[H]
\caption{Closed-loop results on the non-unate rule family and the wide branching graph family (depth 4, fan-in up to 4, max fan-out 10).}
\label{tab:stress}
\centering
\scriptsize
\resizebox{\linewidth}{!}{\begin{tabular}{lllccccccc}
\toprule
Family & Reader & Provenance & Struct.\ cov. & Stale cov. & Residual & Local corr. & Cond.\ cert. & Probe err./ep. & Re-deriv./ep. \\
\midrule
non-unate & Luna & No repair & -- & -- & -- & -- & -- & 1.66 [1.44, 1.88] & 0.0 \\
 &  & Rule-based but-for & 0.69 & 0.95 [0.91, 0.98] & 0.058 & 0.993 & 0.909 & 0.22 [0.10, 0.36] & 7.0 \\
 &  & Matched random & 0.90 & 0.95 [0.91, 0.98] & 0.057 & 0.993 & 0.890 & 0.40 [0.20, 0.64] & 9.0 \\
 &  & Value-agreement & 0.40 & 0.42 [0.36, 0.48] & 0.532 & 0.991 & 0.379 & 1.46 [1.20, 1.70] & 4.0 \\
 &  & Minimal certificate & 0.77 & 0.99 [0.98, 1.00] & 0.011 & 0.993 & 1.000 & 0.08 [0.02, 0.16] & 7.9 \\
 &  & LLM material-support & 0.84 & 0.99 [0.97, 1.00] & 0.014 & 0.992 & 0.998 & 0.14 [0.06, 0.24] & 8.6 \\
 &  & Oracle structural & 1.00 & 1.00 [1.00, 1.00] & 0.000 & 0.993 & 1.000 & 0.12 [0.04, 0.22] & 10.2 \\
\midrule
 & DeepSeek & No repair & -- & -- & -- & -- & -- & 1.58 [1.36, 1.80] & 0.0 \\
 &  & Rule-based but-for & 0.69 & 0.96 [0.93, 0.99] & 0.046 & 1.000 & 0.910 & 0.14 [0.04, 0.24] & 7.0 \\
 &  & Matched random & 0.91 & 0.96 [0.93, 0.99] & 0.037 & 1.000 & 0.892 & 0.20 [0.04, 0.40] & 9.1 \\
 &  & Value-agreement & 0.39 & 0.41 [0.35, 0.47] & 0.536 & 1.000 & 0.372 & 1.38 [1.12, 1.64] & 4.0 \\
 &  & Minimal certificate & 0.77 & 1.00 [1.00, 1.00] & 0.000 & 1.000 & 1.000 & 0.00 [0.00, 0.00] & 7.9 \\
 &  & LLM material-support & 0.85 & 1.00 [0.99, 1.00] & 0.004 & 1.000 & 0.998 & 0.02 [0.00, 0.06] & 8.7 \\
 &  & Oracle structural & 1.00 & 1.00 [1.00, 1.00] & 0.000 & 1.000 & 1.000 & 0.00 [0.00, 0.00] & 10.2 \\
\midrule
wide (depth 4) & Luna & No repair & -- & -- & -- & -- & -- & 1.44 [1.02, 1.90] & 0.0 \\
 &  & Rule-based but-for & 0.16 & 0.73 [0.62, 0.83] & 0.197 & 0.994 & 0.681 & 0.46 [0.20, 0.78] & 4.1 \\
 &  & Matched random & 0.53 & 0.60 [0.42, 0.77] & 0.267 & 0.992 & 0.625 & 0.64 [0.28, 1.04] & 13.2 \\
 &  & Value-agreement & 0.45 & 1.00 [0.99, 1.00] & 0.000 & 0.995 & 1.000 & 0.06 [0.00, 0.16] & 11.5 \\
 &  & Minimal certificate & 0.33 & 0.98 [0.96, 1.00] & 0.005 & 0.993 & 1.000 & 0.08 [0.00, 0.18] & 8.4 \\
 &  & LLM material-support & 0.49 & 1.00 [1.00, 1.00] & 0.000 & 0.995 & 0.998 & 0.08 [0.00, 0.22] & 12.5 \\
 &  & Oracle structural & 1.00 & 1.00 [1.00, 1.00] & 0.000 & 0.994 & 1.000 & 0.02 [0.00, 0.06] & 25.8 \\
\midrule
 & DeepSeek & No repair & -- & -- & -- & -- & -- & 1.40 [0.94, 1.88] & 0.0 \\
 &  & Rule-based but-for & 0.16 & 0.72 [0.61, 0.83] & 0.193 & 1.000 & 0.684 & 0.40 [0.16, 0.70] & 4.1 \\
 &  & Matched random & 0.52 & 0.59 [0.42, 0.76] & 0.277 & 1.000 & 0.622 & 0.60 [0.28, 1.00] & 13.0 \\
 &  & Value-agreement & 0.45 & 1.00 [1.00, 1.00] & 0.000 & 1.000 & 1.000 & 0.00 [0.00, 0.00] & 11.7 \\
 &  & Minimal certificate & 0.33 & 1.00 [1.00, 1.00] & 0.000 & 1.000 & 1.000 & 0.00 [0.00, 0.00] & 8.6 \\
 &  & LLM material-support & 0.49 & 1.00 [1.00, 1.00] & 0.000 & 1.000 & 0.998 & 0.00 [0.00, 0.00] & 12.6 \\
 &  & Oracle structural & 1.00 & 1.00 [1.00, 1.00] & 0.000 & 1.000 & 1.000 & 0.00 [0.00, 0.00] & 25.8 \\
\bottomrule
\end{tabular}
}
\end{table}

\FloatBarrier
\section{Attribution, Repair, and Runtime Cost Accounting}
\label{app:cost}

Table~\ref{tab:cost} reports the per-episode API calls, input/output tokens, and wall-clock latency for both attribution and repair pooled across all five depths $D\in\{1,\dots,5\}$ ($250$ episode-depth runs per reader; hence oracle closure averages $2{,}659/250=10.6$ re-derivations per episode here versus $1{,}625/150=10.8$ on the $D\in\{1,3,5\}$ subset in Table~\ref{tab:closed_loop}). Value-agreement, minimal-certificate, and LLM provenance reduce repair calls, tokens, and latency, with the method-specific costs shown separately. Running an external LLM attributor at every memory formation incurs $13.9$--$15.5$ attribution calls per episode. With thinking enabled, \texttt{deepseek-flash} generates roughly $88\text{k}$--$89\text{k}$ reasoning tokens per episode for attribution; with low reasoning effort, \texttt{gpt-5.6-luna} uses $12.5\text{k}$ total attribution tokens per episode. Recording provenance edges without an LLM call adds $1.8\,\text{ms}$ of CPU time and $384\text{ bytes}$ per episode. These measurements cover symbolic edge logging and separate LLM attribution.

\begin{table}[!htbp]
\caption{Per-episode attribution and repair cost pooled across all five depths $D\in\{1,\dots,5\}$ ($250$ episode-depth runs per reader).}
\label{tab:cost}
\centering
\scriptsize
\resizebox{\linewidth}{!}{\begin{tabular}{llcccccc}
\toprule
Reader & Provenance & Attr.\ calls & Attr.\ tokens (in / out) & Attr.\ latency (s) & Re-deriv. & Repair tokens (in / out) & Repair latency (s) \\
\midrule
Luna & Oracle structural & 0.0 & 0 / 0 & 0 & 10.6 & 3,965 / 745 & 20 \\
 & Value-agreement & 0.0 & 0 / 0 & 0 & 5.6 & 2,062 / 412 & 11 \\
 & Minimal certificate & 0.0 & 0 / 0 & 0 & 4.6 & 1,670 / 336 & 9 \\
 & LLM (deepseek-flash, thinking) & 13.9 & 12,726 / 89,457 & 321 & 5.9 & 2,171 / 435 & 12 \\
 & LLM (gpt-5.6-luna, low) & 15.3 & 8,860 / 3,608 & 58 & 7.3 & 2,705 / 541 & 14 \\
\midrule
DeepSeek & Oracle structural & 0.0 & 0 / 0 & 0 & 10.6 & 3,968 / 1,761 & 15 \\
 & Value-agreement & 0.0 & 0 / 0 & 0 & 5.8 & 2,103 / 974 & 8 \\
 & Minimal certificate & 0.0 & 0 / 0 & 0 & 4.7 & 1,709 / 789 & 7 \\
 & LLM (deepseek-flash, thinking) & 14.0 & 12,726 / 87,991 & 315 & 6.0 & 2,179 / 1,013 & 8 \\
 & LLM (gpt-5.6-luna, low) & 15.5 & 8,955 / 3,624 & 58 & 7.5 & 2,762 / 1,243 & 12 \\
\bottomrule
\end{tabular}
}
\end{table}

\section{Failure Taxonomy and Stronger-Model Checks}
\label{app:failures}

To audit how closely the empirical runs satisfy Assumptions (i) and (ii) of Proposition~\ref{prop:certificate}, Table~\ref{tab:failures} separates \textbf{local correctness} (the share of formations where the reader LLM computed $y_v=f_v((y_p)_{p\in\mathrm{Pa}(v)})$ correctly from its stored parents) from \textbf{conditional certificate containment} (the share of locally correct formations whose recorded edges contain a certificate of $y_v$). Earlier versions of our diagnostic combined both criteria into a single joint rate, which made symbolic value-agreement, minimal-certificate, and oracle provenance on Luna appear below $1.00$ ($0.989$--$0.993$) solely because Luna made $20$--$21$ local calculation errors per closed-loop run ($98.94\%$--$99.29\%$ local correctness). Conditioning on locally correct formations shows that both value-agreement ($2{,}029/2{,}029$ on Luna, $2{,}064/2{,}064$ on DeepSeek) and minimal-certificate provenance ($1{,}865/1{,}865$ on Luna, $1{,}904/1{,}904$ on DeepSeek) achieve $100\%$ conditional certificate containment, matching Proposition~\ref{prop:certificate} with zero certificate omissions.

For LLM material-support provenance (\texttt{deepseek-flash}), conditional certificate containment is $99.95\%$ ($2{,}069/2{,}070$, out of $2{,}091$ total formations) on Luna and $99.81\%$ ($2{,}092/2{,}096$) on DeepSeek. Because $5$ locally correct formations in the closed-loop depth sweep ($1$ on Luna, $4$ on DeepSeek; $3$ in the replay sweep) omitted a certificate member, Proposition~\ref{prop:certificate}'s worst-case guarantee does not apply unconditionally to LLM provenance. Whereas certificate containment is a sufficient condition guaranteeing repair across \emph{all} possible update sequences, in the observed correction sequences those $5$ certificate omissions produced $0$ additional missed stale pairs during repair. All $16$ missed stale pairs ($11$ wrong when written, $5$ inherited from a wrong ancestor) and all $9$ residual stale states ($4$ wrong when written, $5$ inherited) on Luna instead trace to the $21$ local reader calculation errors violating Assumption (i).

When the unique failing reader prompts are re-queried at low reasoning effort, \texttt{gpt-5.6-luna} answers $87\%$ ($26/30$) correctly; when queried at high reasoning effort or with a stronger model (\texttt{gpt-5.6-terra}), accuracy on those prompts reaches $100\%$ ($10/10$) in both cases. Likewise, on the three correctly stored replay states where \texttt{deepseek-flash} attribution omitted a certificate member, re-running attribution with \texttt{gpt-5.6-terra} recovers a valid certificate in $100\%$ ($3/3$) of cases. These results distinguish local reader errors from certificate omissions: all observed missed stale pairs trace to the former, while the recorded omissions produce no additional misses on the evaluated sequences.

\begin{table}[!htbp]
\caption{Root-cause breakdown of unreached stale states in the depth sweep and stronger-model verification checks.}
\label{tab:failures}
\centering
\small
\resizebox{\linewidth}{!}{\begin{tabular}{lcc}
\toprule
Diagnostic / failure source (closed-loop \& replay depth sweep) & Luna reader & DeepSeek reader \\
\midrule
Minimal certificate: local correctness (correct formations / total) & 0.9894 (1,865 / 1,885) & 1.0000 (1,904 / 1,904) \\
Minimal certificate: conditional certificate containment (given correct) & 1.0000 (1,865 / 1,865) & 1.0000 (1,904 / 1,904) \\
Value-agreement: local correctness (correct formations / total) & 0.9898 (2,029 / 2,050) & 1.0000 (2,064 / 2,064) \\
Value-agreement: conditional certificate containment (given correct) & 1.0000 (2,029 / 2,029) & 1.0000 (2,064 / 2,064) \\
LLM provenance: local correctness (correct formations / total) & 0.9900 (2,070 / 2,091) & 1.0000 (2,096 / 2,096) \\
LLM provenance: conditional certificate containment (given correct) & 0.9995 (2,069 / 2,070) & 0.9981 (2,092 / 2,096) \\
LLM provenance: formations omitting a certificate (given correct) & 1 & 4 \\
LLM provenance: missed stale pairs from certificate omissions & 0 & 0 \\
\midrule
missed stale pairs: value wrong when written (local reader error) & 11 & 0 \\
missed stale pairs: inherited wrong ancestor & 5 & 0 \\
residual states: value wrong when written (local reader error) & 4 & 0 \\
residual states: inherited wrong ancestor & 5 & 0 \\
\midrule
Re-asking the failing prompts (unique prompts) & \multicolumn{2}{c}{fraction answered correctly} \\
gpt-5.6-luna, low effort, three fresh samples & \multicolumn{2}{c}{0.87 (n=30)} \\
gpt-5.6-luna, high effort & \multicolumn{2}{c}{1.00 (n=10)} \\
gpt-5.6-terra, low effort & \multicolumn{2}{c}{1.00 (n=10)} \\
Re-attribution of correct states lacking a certificate, gpt-5.6-terra & \multicolumn{2}{c}{1.00 (n=3)} \\
\bottomrule
\end{tabular}
}
\end{table}

\end{document}